%% file: main.tex
\documentclass[runningheads]{llncs}

\usepackage{eccv}

\usepackage{eccvabbrv}

\usepackage{graphicx}
\usepackage{booktabs}
\usepackage{bbding} 
\usepackage[table]{xcolor} 
\usepackage{subcaption} 
\usepackage{caption}
\usepackage{multirow}
\usepackage{enumitem}

\usepackage[accsupp]{axessibility}  

\usepackage[pagebackref,breaklinks,colorlinks,citecolor=eccvblue]{hyperref}

\usepackage{orcidlink}
\usepackage{setspace}

\definecolor{mygreen}{HTML}{B9FFBF}
\definecolor{mylightgreen}{HTML}{D9F0D9}

\usepackage[most]{tcolorbox}
\tcbuselibrary{listings,breakable}

\newtcblisting{jsonbox}{
  breakable,
  listing only,
  colback=gray!12,
  colframe=black,
  boxrule=0.8pt,
  arc=2pt,
  left=8pt,
  right=8pt,
  top=8pt,
  bottom=8pt,
  listing options={
    basicstyle=\ttfamily\small,
    columns=fullflexible,
    keepspaces=true,
    breaklines=true,
    showstringspaces=false
  }
}

\usepackage{tabularx}
\usepackage{etoolbox}
\makeatletter
\newcommand{\tableofcontentsonthispage}{%
  \begingroup
  \let\clearpage\relax
  \let\cleardoublepage\relax
  \tableofcontents
  \endgroup
}
\makeatother

\begin{document}

\title{Count Anything at Any Granularity} 

\titlerunning{Abbreviated paper title}

\author{Chang Liu\inst{1,2} \and
Haoning Wu\inst{1} \and
Weidi Xie\inst{1}}

\authorrunning{C.~Liu et al.}

\institute{School of Artificial Intelligence, Shanghai Jiao Tong University, China \and
CMIC, Shanghai Jiao Tong University, China}

\maketitle

\input{sec/0_abstract}
\input{sec/1_introduction}

\input{sec/2_relatedworks}

\input{sec/3_taskdefinition}
\input{sec/4_dataset}

\input{sec/5_experiments}
\input{sec/6_conclusion}

%
%
\clearpage
\bibliographystyle{splncs04}
\bibliography{main}

\input{sec/X_supp}

\end{document}

%% file: sec/0_abstract.tex
\begin{abstract}
Open-world object counting remains brittle: despite rapid advances in vision-language models~(VLMs), reliably counting the objects a user intends is far from solved. 
We argue that a central reason is that counting granularity is left implicit; users may refer to a specific identity, an attribute, an instance type, a category, or an abstract concept, yet most methods treat ``what to count'' as a single, category-level matching problem. 
In this work, we redefine open-world counting as multi-grained counting, where visual exemplars specify target appearance and fine-grained text~(with optional negative prompts) specifies the intended semantic granularity across five explicit levels. 
Making granularity explicit, however, exposes a critical data bottleneck: existing counting datasets lack the multi-category scenes, controlled distractors, and instance-level annotations needed to verify fine-grained prompt semantics. 
To address this, we propose the first fully automatic data-scaling pipeline that integrates controllable 3D synthesis with consistent image editing and VLM-based filtering, and use it to construct \textbf{KubriCount}, the largest and most comprehensively annotated counting dataset to date, supporting both training and multi-grained evaluation. 
Systematic benchmarking reveals that both multimodal large language models and specialist counting models exhibit severe prompt-following failures under fine-grained distinctions. Motivated by these findings, we train \textbf{HieraCount}, a multi-grained counting model that jointly leverages text and visual exemplars as complementary target specifications. 
HieraCount substantially improves multi-grained counting accuracy and generalizes robustly to challenging real-world scenarios.
The project page is available \href{https://verg-avesta.github.io/KubriCount/}{here}.
\keywords{Multi-Grained Counting \and Data Scaling \and Benchmark}
\end{abstract}


%% file: sec/1_introduction.tex
\section{Introduction}
The ability to perceive numerosity, ranging from rapid subitizing to deliberate counting, is a fundamental cognitive skill present even in early human infancy~\cite{Kaufman49}. 
In computer vision, however, a striking paradox exists: while large-scale foundation models~\cite{Qwen2.5-VL, Qwen3-VL, InternVL2.5, kimivl} have driven remarkable progress on complex multimodal reasoning tasks~({\em e.g.}, visual-question answering), reliably counting \emph{the objects a user intends} in open-world images remains brittle. 
Despite rapid advances in open-world detection~\cite{groundingdino, jiang2024t,rexomni} and segmentation~\cite{sam, sam2, sam3}, object counting has advanced at a noticeably slower pace, remaining far from a reliable and unified formulation.

We argue that a central reason is conceptual rather than architectural: \textbf{counting has rarely been posed as a strictly well-defined, user-controllable problem}. 
The literature has progressed from \emph{class-specific} counting for predefined categories ({\em e.g.}, pedestrians, cells)~\cite{arteta2016counting,mundhenk2016large,xie2018microscopy}, to \emph{class-agnostic} exemplar-based counting~\cite{lu2018class,countr,loca}, and more recently to \emph{open-world counting} enabled by vision-language pretraining~\cite{clip}, where targets are specified by free-form text or exemplars~\cite{countgd,countx,pelhan2024dave,kang2024vlcounter}.
Yet most existing formulations define ``what to count'' primarily at the \emph{category} level. Real-world scenes, by contrast, exhibit a multiple hierarchies, {\em i.e.}, users may mean an object \emph{identity} (a specific item), an \emph{attribute} ({\em e.g.}, red cars), an \emph{instance type} ({\em e.g.}, sedan vs.\ SUV), a \emph{category} (cars), or an abstract \emph{concept} (things used for driving).
When this hierarchy is left implicit, the matching criterion becomes ill-specified: in the presence of distractors, models can satisfy a query in unintended ways, often defaulting to visually dominant or repetitive groups, leading to poor prompt-following behavior~\cite{countgd,pelhan2024geco}.

\begin{figure*}[t]
  \centering
  \includegraphics[width=\textwidth]{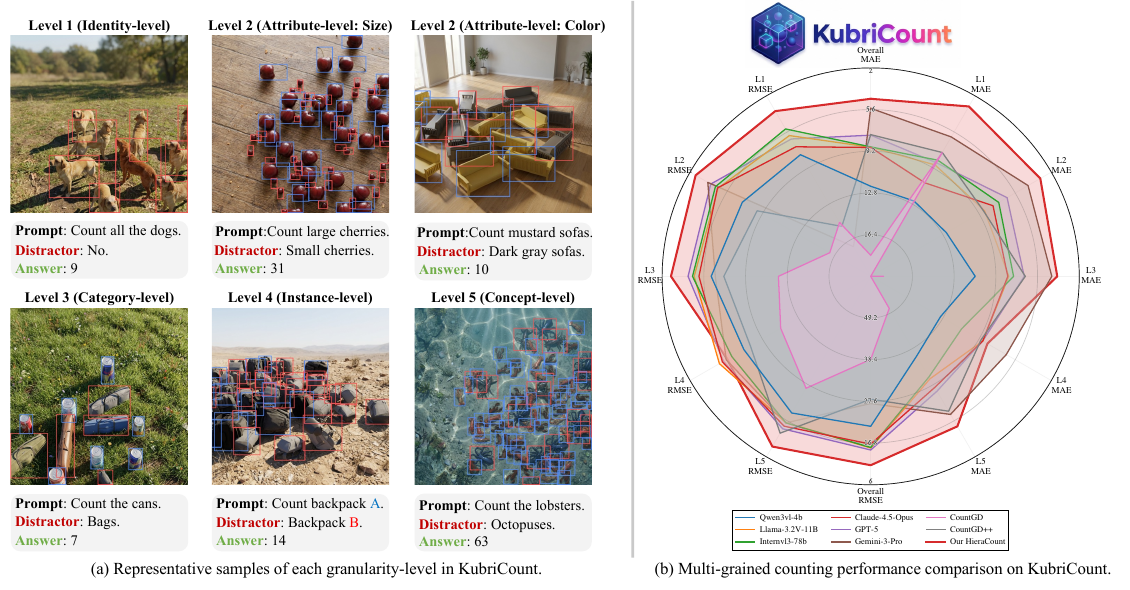}
  \vspace{-16pt}
  \caption{
    \textbf{Multi-grained counting benchmark~(KubriCount) and model evaluation.
  }
    \textbf{Left:} KubriCount examples across five granularity levels~(L1--L5), illustrating how prompts specify different counting scopes.
    \textbf{Right:} Multi-grained counting performance comparison of representative MLLMs and expert models across levels.}
  \vspace{-16pt}
  \label{fig:teaser}
\end{figure*}

In this work, we take a step toward a more robust and controllable formulation by redefining open-world counting as \textbf{multi-grained counting}, where the visual exemplars specify the target appearance, while fine-grained text specifies the intended semantic granularity (with optional negative prompts for further disambiguation). 
Concretely, we decompose user intent into \textbf{five semantic levels}: \emph{identity, attribute, category, instance}, and \emph{concept}.
This turns open-world counting into a \emph{verifiable prompt-following} problem:
the question is not to output a plausible number, but to count the correct set under an explicit prompts.

Making granularity explicit, however, exposes a key bottleneck: \textbf{the lack of scalable, high-quality data that can verify fine-grained prompt semantics}.
Multi-grained evaluation requires multi-category scenes, controlled hard negatives, and instance-level annotations that specify which objects should (and should not) be counted at each level. 
Most existing counting datasets~\cite{fsc147,omnicount} were designed for simpler regimes; 
they are typically small-scale and often single-category, limiting their ability to stress-test fine-grained distinctions. 
This scarcity is driven by two practical constraints: collecting high-density, multi-category images at scale is difficult, and dense manual annotation is prohibitively expensive. As a result, counting data has scaled slowly, and progress on robust, controllable counting has lagged behind other open-world perception tasks.

To address this gap, we propose the \textbf{first fully automatic pipeline for scaling counting data}, integrating controllable 3D synthesis with consistent image editing. We curate a diverse pool of 3D assets and use the Kubric engine~\cite{kubric} to synthesize multi-object image prototypes with exact instance-level metadata. To narrow the sim-to-real gap while preserving annotations, we apply consistent image editing~\cite{nanobananapro} and then use VLM-based filtering~\cite{gemini3} to remove samples with semantic or geometric inconsistencies. Built on this pipeline, we construct \textbf{KubriCount}, a large-scale benchmark featuring controlled distractors and supervision aligned with all five semantic granularity levels, designed to advance prompt-following in multi-grained counting.

Using KubriCount, we conduct a comprehensive evaluation of representative multimodal large language models (MLLMs) and specialist counting models. We find that both families exhibit systematic prompt-following failures under multi-category distractors and fine-grained distinctions, indicating that open-world counting remains far from robust. 
Motivated by these findings, we train \textbf{HieraCount}, a multi-grained counting model on KubriCount. 
By jointly accepting text prompts and visual exemplars as complementary specifications of the target set, HieraCount substantially improves multi-grained counting and generalizes well to challenging real-world scenarios.

To summarize, we make the following contributions in this paper:
(i) we define a \textbf{multi-grained counting} task, rendering counting granularity explicit and verifiable;
(ii) we propose the first fully automatic pipeline for scaling counting data and construct \textbf{KubriCount}, the largest and most comprehensively annotated object counting dataset to date, supporting both training and multi-grained evaluation;
(iii) we develop \textbf{HieraCount}, a multi-grained counting model trained with granularity-aware prompts and complementary text/exemplar prompting; 
and (iv) we conduct extensive evaluations on MLLMs and counting expert models, demonstrating that HieraCount significantly advances prompt-following counting with robust real-world generalization.

%% file: sec/2_relatedworks.tex
\section{Related Work}

\noindent \textbf{Counting models.}
Early counting systems typically employ closed-set detectors~\cite{he2017mask, lin2017focal} to derive counts directly from detected bounding boxes. 
For highly dense scenes, density-map regression~\cite{arteta2014interactive, arteta2016counting, kong2006viewpoint, lempitsky2010learning, marana1997estimation, xie2018microscopy} has emerged as a more accurate and robust alternative~\cite{desai2011discriminative,barinova2012detection,carpk,nguyen2022few,loca}.
Building on this paradigm, exemplar-based regressors like CounTX~\cite{countx} and CounTR~\cite{countr} predict a density map conditioned on visual exemplars to estimate the final count; however, they inherently lack explicit instance localization and rely heavily on Gaussian surrogates. 
Recently, the integration of vision-language foundation models~\cite{clip} and open-world detectors~\cite{groundingdino} has enabled methods such as GroundingREC~\cite{groundingrec}, CountGD~\cite{countgd}, and CountGD++~\cite{countgd++} to achieve superior localization and counting performance.
Concurrently, modern multimodal large language models~(MLLMs)~({\em e.g.}, Qwen-VL~\cite{Qwen2.5-VL, Qwen3-VL} and Gemini~\cite{gemini2.5,gemini3}) have exhibited emergent counting capabilities through direct prompting, though their reliability in crowded scenarios remains an open research question.

\vspace{2pt}
\noindent \textbf{Counting prompts.}
Building upon the class-agnostic paradigm introduced in~\cite{lu2018class}, early prompt-based methods~\cite{loca,countr,lu2018class,nguyen2022few,fsc147,shi2022represent,you2023few,lin2022scale,gong2022class,yang2021class} are predominantly driven by visual exemplars.
Subsequent works~\cite{groundingrec,kang2024vlcounter,xu2023zero,jiang2023clip,countx} explore expanding this scope by incorporating text prompts, serving as either substitutes for or complements to visual crops. 
For instance, CountGD~\cite{countgd} seamlessly integrates both modalities into a unified framework.
Furthermore, CountGD++~\cite{countgd++} pioneers the use of negative prompts to enable fine-grained control, highlighting the necessity of explicit disambiguation.

\vspace{2pt}
\noindent \textbf{Counting datasets.}
Early benchmarks are heavily constrained to single-category or domain-specific scenarios~({\em e.g.}, ShanghaiTech~\cite{shanghaitech} and CARPK~\cite{carpk}).
While VQA-style counting datasets such as TallyQA~\cite{tallyqa}, CountBench~\cite{countbench}, and pixmo-count~\cite{deitke2025molmo} offer question-answer supervision, they critically lack precise instance-level spatial annotations. 
Conversely, exemplar-based class-agnostic datasets, notably FSC-147~\cite{fsc147} and OmniCount-191~\cite{omnicount}, provide point labels and prompts; however, their images frequently feature only a single dominant category. 
This homogeneity fails to penalize models that rely on superficial texture matching rather than genuine semantic understanding. 
Although recent datasets like PrACo~\cite{praco} and PairTally~\cite{pairtally} introduce hard negatives to probe fine-grained prompt adherence, their construction remains fundamentally bottlenecked by manual annotation. 
Consequently, they struggle to scale and often suffer from collection artifacts or category bias, thereby motivating the automated, scalable data generation pipeline proposed in this work.

%% file: sec/3_taskdefinition.tex
\section{Multi-Grained Counting}
This section presents a unified formulation and model for multi-grained counting.
We first formalize the \textbf{multi-grained counting} task~(\cref{sec:task_definition}), then introduce \textbf{HieraCount}, a multi-grained counting model that enumerates the prompt-specified target set at an explicit granularity~(\cref{sec:model_arch} and \cref{sec:model_prompt}).

\subsection{Problem Formulation}
\label{sec:task_definition}

\begin{figure*}[t]
  \centering
  \includegraphics[width=\textwidth]{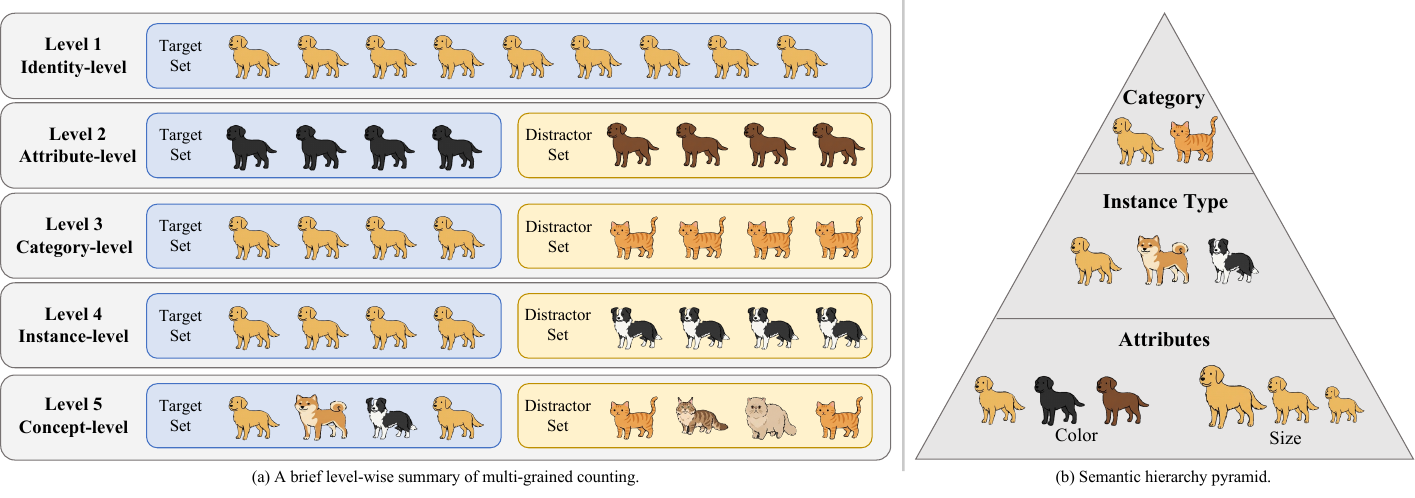}
  \vspace{-16pt}
  \caption{
    \textbf{Multi-grained counting levels and semantic hierarchy.}
    \textbf{Left:} Schematic examples of Levels~1--5, showing the target set \(\mathcal{S}^+\) (blue) and the distractor set \(\mathcal{S}^-\) (yellow).
    \textbf{Right:} The semantic hierarchy underlying our task, from \textbf{category} to \textbf{instance type} to \textbf{attributes} (color/size).
  }
  \label{fig:level_hierarchy}
  \vspace{-12pt}
\end{figure*}

In this paper, we consider a \textbf{multi-grained counting} task whose granularity specification is explicit and controllable.
Specifically, given an image~(\(I\)) with object set \(\mathcal{O}=\{o_1,\dots,o_N\}\), each object~(\(o_i\)) is associated with a category label~(\(c_i\)), an instance type~(\(t_i\)), and an attribute tuple \(\mathbf{a}_i=(\sigma_i,\gamma_i)\) indicating size and color, \textit{i.e.}, \(o_i=(c_i,t_i,\mathbf{a}_i)\). 
Naturally, it adopts a semantic hierarchy \textbf{category \(\supset\) instance  \(\supset\) attribute}. 
A query (specified by text and/or visual exemplars) can be defined as the finest semantic factor in the hierarchy required to distinguish a target subset \(\mathcal{S}^+\subseteq\mathcal{O}\) from an optional distractor subset \(\mathcal{S}^-\subseteq\mathcal{O}\), with \(\mathcal{S}^+\cap\mathcal{S}^-=\emptyset\). 
\textbf{Our goal is therefore to predict the count \(y=|\mathcal{S}^+|\).} 

We instantiate five levels by defining \(\mathcal{S}^+\) and \(\mathcal{S}^-\) such that they only differ along one factor in the hierarchy (category / instance type / attributes).

\vspace{-3pt}
\begin{enumerate}[label=\roman*).]
    \setlength \itemsep{5pt}
    \item \textbf{Level 1 (identity-level).}
    The image contains objects with a single category~(\(c\)), instance type~(\(t\)) and attributes~(\(\mathbf{a}\)); we set \(\mathcal{S}^+=\mathcal{O}\) and \(\mathcal{S}^-=\emptyset\).
    This level mirrors FSC-147~\cite{fsc147} and tests counting all instances in the image.

    \item \textbf{Level 2 (attribute-level).}
    All objects share the same category \(c\) and instance type \(t\), while \(\mathcal{S}^+\) and \(\mathcal{S}^-\) differ in \emph{exactly one} attribute:
    either size-mode (\(\sigma^+\neq\sigma^-\) with a fixed \(\gamma\)), or color-mode (\(\gamma^+\neq\gamma^-\) with a fixed \(\sigma\)).
    The goal is to count the target attribute variant and exclude the other.

    \item \textbf{Level 3 (category-level).}
    \(\mathcal{S}^+\) and \(\mathcal{S}^-\) belong to different categories (\(c^+\neq c^-\)), and each group is restricted to a single instance type (\(t^+\) and \(t^-\)).
    This isolates category-level discrimination from intra-category variation.

    \item \textbf{Level 4 (instance-level).}
    \(\mathcal{S}^+\) and \(\mathcal{S}^-\) share the same category~(\(c\)) but have different instance types (\(t^+\neq t^-\)).
    This requires fine-grained within-category discrimination to separate near-neighbor distractors.

    \item \textbf{Level 5 (concept-level).}
    \(\mathcal{S}^+\) and \(\mathcal{S}^-\) belong to different categories (\(c^+\neq c^-\)), and each group spans at least two instance types to induce large intra-category variation:
    \(|\{t_i: o_i\in\mathcal{S}^+\}|\ge 2\) and \(|\{t_j: o_j\in\mathcal{S}^-\}|\ge 2\).
    This setting stresses robustness beyond single-instance category matching.

\end{enumerate}

We place category-level counting (L3) before instance-level counting (L4) since distinguishing across categories is often easier than separating near-neighbor instance types within a category.
For an intuitive summary of how each level varies category/instance-type/attribute, see~\cref{fig:level_hierarchy}.

\subsection{HieraCount Architecture}
\label{sec:model_arch}

\textbf{HieraCount} is an open-world counting architecture designed for prompt following in cluttered, multi-category scenes with hard distractors. Given a query that induces a target set \(\mathcal{S}^+\) (and optional \(\mathcal{S}^-\)), it takes an image~(\(I\)), an optional set of visual exemplar boxes \(\mathcal{B}=\{b_j\}_{j=1}^{m}\) sampled from instances in \(\mathcal{S}^+\), and an optional text prompt~(\(p\)) specifying the intended granularity, and predicts the count \(\hat y=f(I,\mathcal{B},p)\) by localizing and enumerating target instances.
We train HieraCount on our multi-grained counting dataset, demonstrating the effect of \emph{multi-grained supervision and prompts}.

\vspace{3pt}
\noindent \textbf{Image and text encoders.}
HieraCount inherits the vision-language backbone from GroundingDINO~\cite{groundingdino}, with an image encoder~(\(f^{I}_{\theta}\)) and a text encoder~(\(f^{T}_{\theta}\)).
The image encoder maps \(I\) to multi-scale image features, projected to a common embedding dimension~(\(d\)) to form image tokens \(\mathcal{Z}_I\in\mathbb{R}^{n\times d}\).
The text encoder maps \(p\) to token features \(\mathcal{Z}_p\in\mathbb{R}^{q\times d}\).
We extract exemplar tokens from the image feature maps using RoI-Align over the boxes~(\(\mathcal{B}\)).
This yields visual exemplar tokens \(\mathcal{Z}_v\in\mathbb{R}^{m\times d}\) that share the same feature space, enabling seamless multimodal fusion and supporting a variable number of exemplars.

\vspace{3pt}
\noindent\textbf{Prompt-image fusion.}
A feature enhancer~(\(f_{\phi}\)) fuses the visual and text prompt tokens, and propagates them to the image features via attention.
Concretely, it produces fused prompt tokens~(\(\mathcal{Z}_{v,p}\)) and enhanced image tokens~(\(\tilde{\mathcal{Z}}_I\)):
\[
(\mathcal{Z}_{v,p}, \tilde{\mathcal{Z}}_I) = f_{\phi}(\mathcal{Z}_v, \mathcal{Z}_p, \mathcal{Z}_I).
\]
Intuitively, self-attention over \((\mathcal{Z}_v, \mathcal{Z}_p)\) allows the model to combine complementary cues (visual appearances and language semantics), while cross-attention aligns the fused prompt representation with relevant regions in the image.

\vspace{3pt}
\noindent \textbf{Query selection, decoding, and counting.}
From \(\tilde{\mathcal{Z}}_I\), we select the top-\(k\) image tokens that are most relevant to \(\mathcal{Z}_{v,p}\) (by similarity) as cross-modality queries, and feed them to a cross-modality decoder~(\(f_{\psi}\)).
The decoder outputs a set of \(k\) candidate instances with localization predictions and prompt-conditioned confidence scores.
We then threshold the confidence scores to obtain final detections, and compute the count as the number of retained instances.

\vspace{3pt}
\noindent \textbf{Training objective and implementation.}
Following~\cite{countgd}, we train the visual exemplar projection layers, the feature enhancer~(\(f_{\phi}\)), and the decoder~(\(f_{\psi}\)), while keeping \(f^{I}_{\theta}\) and \(f^{T}_{\theta}\) frozen.
Training uses bipartite Hungarian matching between predicted queries and ground-truth instances, with an additional ``no-object'' label for unmatched queries.
The overall loss~($\mathcal{L}$) is a weighted sum of a localization loss~($\mathcal{L}_{\text{loc}}$) and a classification loss~($\mathcal{L}_{\text{cls}}$), represented as:
\[
    \mathcal{L} = \lambda_{\text{loc}} \cdot \mathcal{L}_{\text{loc}} + \lambda_{\text{cls}} \cdot \mathcal{L}_{\text{cls}},
\]
where \(\mathcal{L}_{\text{loc}}\) regresses instance locations ({\em e.g.}, centers) and \(\mathcal{L}_{\text{cls}}\) supervises the prompt-conditioned confidence scores (using the same objectives as~\cite{countgd}).
We train HieraCount for two epochs on the KubriCount training split, and mix in FSC-147~\cite{fsc147} for stability, while keeping the other hyperparameters consistent with~\cite{countgd} so that improvements reflect the effect of data and prompts.

\subsection{Granularity-aware Prompts}
\label{sec:model_prompt}
\noindent\textbf{Hybrid queries with explicit granularity.}
Multi-grained counting provides supervision via the semantic hierarchy \((c_i,t_i,\mathbf{a}_i)\).
To align HieraCount with this hierarchy, we construct \emph{one training item per object group} in an image by pairing:
(i) a small set of exemplar boxes \(\mathcal{B}\) sampled from the target group \(\mathcal{S}^+\), and
(ii) a level-dependent text phrase that explicitly reflects the intended granularity.
This replaces the common practice of pairing an instance-level exemplar with a coarse category name, and encourages consistent semantics between exemplar matching and language guidance.

\vspace{3pt}
\noindent \textbf{Multi-phrase captions with negatives.}
During training, we adopt a multi-phrase caption format following GroundingDINO: each item includes the positive target phrase and a few sampled negative phrases (distractor descriptions) to encourage discrimination under multi-category clutter.
These negatives are used as training supervision signals rather than as a test-time interface.
At inference time, we evaluate HieraCount using positive-only text and visual prompting, so gains reflect improved positive matching and prompt faithfulness.

\vspace{3pt}
\noindent \textbf{Discussion.}
Overall, HieraCount keeps the original architecture fixed, but trains it with multi-grained, granularity-aware prompts.
This design leverages the complementarity of visual exemplars (appearance grounding) and fine-grained language (semantic granularity specification), matching the goal of multi-grained counting: counting the \emph{prompt-intended} target set at an explicit granularity.

%% file: sec/4_dataset.tex
\section{KubriCount: Data Scaling Pipeline and Benchmark}
High-quality counting data remains the main bottleneck for robust, prompt-following counting.
We present a \textbf{fully automatic data scaling pipeline} that generates scenes with high category diversity, controllable composition, and precise instance-level annotations. Using this pipeline, we build \textbf{KubriCount}, a large-scale benchmark for \textbf{multi-grained} counting with controlled distractors and explicit granularity.

\begin{figure*}[t]
  \centering
  \includegraphics[width=\textwidth]{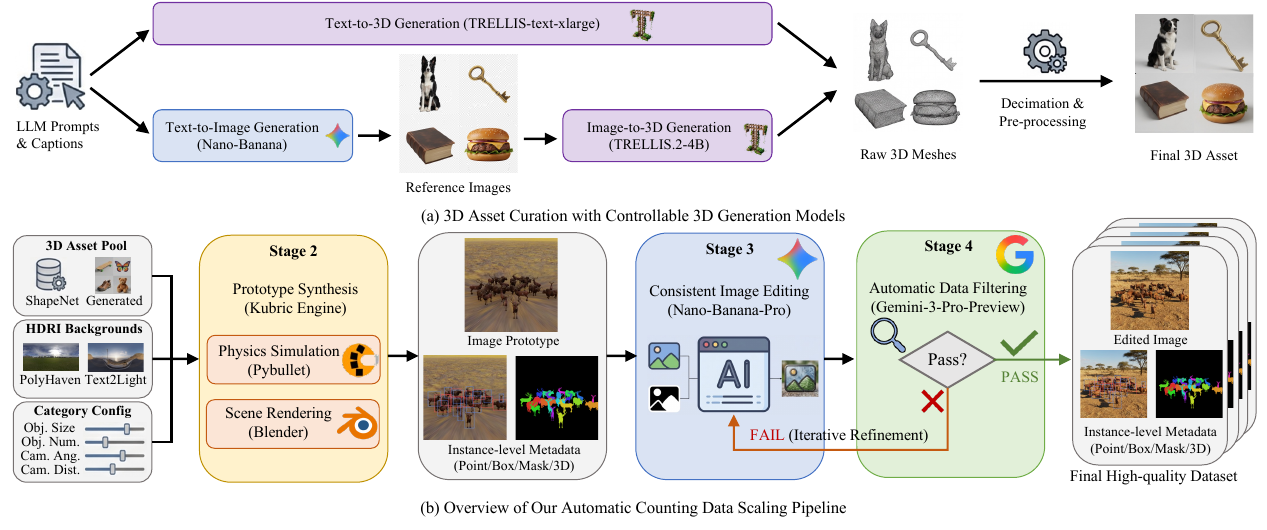}
  \vspace{-18pt}
  \caption{
    \textbf{Overview of our automatic counting data scaling pipeline.}
    We curate and generate 3D assets, synthesize labeled prototypes with Kubric, apply mask-conditioned consistent editing to improve realism, and use VLM-based filtering (with an edit-filter loop) to ensure label consistency.
  }
  \vspace{-16pt}
  \label{fig:pipeline}
\end{figure*}

\subsection{Automatic Data Scaling Pipeline}
\label{sec:auto_pipeline}

As illustrated in \cref{fig:pipeline}, we generate KubriCount in four stages as follows:

\vspace{3pt} \noindent \textbf{Stage-I: 3D asset curation.}
To support multi-grained semantics and prompt construction, we curate a repository of 3D objects with explicit category metadata. We build an asset bank from two sources.
(i) \textbf{Labeled 3D datasets:} ShapeNetCore-v2~\cite{shapenet} (53K assets, 55 categories) with clean taxonomic labels; we avoid repositories with weak or noisy tags ({\em e.g.}, GSO~\cite{gso}, Objaverse~\cite{objaverse,objaversexl}).
(ii) \textbf{Controllable 3D generation:} to better match real-world category long tails~\cite{fsc147,deitke2025molmo}, we generate additional objects with the TRELLIS family~\cite{trellis,trellis2}, using LLM-produced prompts~\cite{gpt5,gemini3}.
We use both text-to-3D~\cite{trellis} and a text-to-image-to-3D route (Nano-Banana~\cite{nanobanana} RGBA cutouts followed by TRELLIS.2-4B~\cite{trellis2} mesh reconstruction).
After preprocessing for Kubric, we obtain $\sim$5K more assets across 102 new categories.
Ultimately, we collect $\sim$\textbf{58K assets} spanning \textbf{157 categories}. 
To ensure diverse scene illumination, we complement these assets with $\sim$5K HDRI environment maps, combining $\sim$500 curated HDRIs from Poly Haven~\cite{polyhaven} with $\sim$4.5K outdoor HDRIs synthesized via Text2Light~\cite{text2light} and filtered through automated sanity checks, {\em e.g.}, failed panoramas.

\vspace{3pt} \noindent \textbf{Stage-II: prototype synthesis.}
Using the curated assets, we synthesize large-scale \emph{image prototypes} via Kubric~\cite{kubric},
that provide exact, controllable instance-level supervision, though they initially exhibit a sim-to-real gap in photorealism.

We enforce strict \textbf{dataset splits} during synthesis.
3D assets are divided into \textbf{Train} (seen categories), \textbf{TestA} (unseen assets within training categories; $\sim$10\% holdout per category), and \textbf{TestB} (unseen categories; $\sim$10\% of total assets).
HDRI backgrounds are split similarly, and both TestA and TestB use only unseen HDRIs.
We also apply \textbf{level-specific scene composition rules} by controlling how assets are selected for the target/distractor groups.
Level~1 samples a single category and instance type; Level~2 keeps them fixed and varies a single attribute; Level~3 varies category; Level~4 varies instance type within the same category; and Level~5 uses two categories with multiple instance types per category to induce larger intra-category variation.
For multi-category Levels~3 and~5, the two categories are chosen from the same \textbf{super-category} (clustered by typical real-world co-occurrence) to form semantically plausible distractors.

Scene generation is \textbf{configuration-driven} with category-specific profiles that set synthesis hyperparameters to guarantee physical plausibility. 
A customized Kubric worker samples assets based on level rules, initializes them within constrained parameter profiles (governing size, density, and camera pose), and executes a rigid-body simulation. The engine renders RGB images alongside pixel-perfect instance masks, 2D/3D bounding boxes, and center points. We utilize both normal and dense configurations to simulate sparse-to-crowded distributions, capping the maximum object count at \textbf{250} per image to comply with Kubric's 256-instance ID limit.

\vspace{3pt} 
\noindent \textbf{Stage-III: consistent image editing.}
We use Nano-Banana-Pro~\cite{nanobananapro} to reduce the sim-to-real gap by refining textures and harmonizing lighting while keeping supervision intact. 
The editing is conditioned on the prototype RGB and masks (instance masks for Levels~1-2; target/distractor/background masks for Levels~3-5) and is constrained to preserve topology, {\em i.e.}, \textbf{no instance is added, removed, merged, or split}. 
The procedure is \textbf{level-aware}: background edits are always allowed, while object edits are restricted when they may change the ground truth~({\em e.g.}, disabling color/texture changes in Level~2 when color specifies the attribute) and kept conservative in Levels~3--5 to maintain target-distractor separability.

\vspace{3pt}
\noindent \textbf{Stage-IV: automatic data filtering.}
Editing can still occasionally break our topology constraints, so we add an automatic filtering step to ensure label fidelity. We use Gemini-3-Pro~\cite{gemini3} as a visual inspector, feeding it the prototype image, instance masks, and the edited result to output a \texttt{PASS}/\texttt{FAIL} verdict. 
We reject samples with (i) viewpoint/layout drift, (ii) instance-count changes (removals, duplications, merges), (iii) target/distractor identity corruption, (iv) background hallucinations, or (v) severe artifacts; minor mask-boundary leakage is allowed if core invariants hold. A single pass filters out $\sim$20\% of edits; we then iteratively re-edit and re-check failed cases, reducing the final rejection rate to $\sim$5\% after three iterations, with the remaining failures discarded.

\input{tables/dataset_comparison}

\subsection{Dataset Statistics}
\label{sec:dataset_statistics}

\cref{tab:dataset_comparison} compares KubriCount with prior counting benchmarks.
KubriCount contains \textbf{$\sim$110K images} with \textbf{$\sim$7M instances} over \textbf{157 categories} in \textbf{16 super-categories}. 
While its image count is second only to TallyQA (which lacks instance-level labels), KubriCount is substantially larger in \emph{annotated instances} and provides dense supervision for counting: per-instance center points, 2D/3D boxes, and pixel-accurate masks. To our knowledge, \textbf{KubriCount is the largest and most comprehensively annotated dataset for visual counting}.

\vspace{3pt} 
\noindent \textbf{Data splits.}
KubriCount spans five counting levels.
Each level includes $\sim$\textbf{20K} training images and two test sets of $\sim$\textbf{1K} images each, totaling $\sim$110K images. To ensure model robustness to varying object densities, the training set mixes normal and dense spatial configurations at a $\sim$\textbf{4:1} ratio, whereas both test splits strictly evaluate on the normal configuration. In total, the benchmark is partitioned into \textbf{Train}~($\sim$100K), \textbf{TestA}~($\sim$5K, images featuring novel assets from seen categories), and \textbf{TestB} ($\sim$5K images featuring entirely novel categories).

\begin{figure*}[t]
    \centering
    \begin{subfigure}[t]{0.40\textwidth}
        \centering
        \includegraphics[
            width=\linewidth,
            height=5.3cm,
            keepaspectratio
        ]{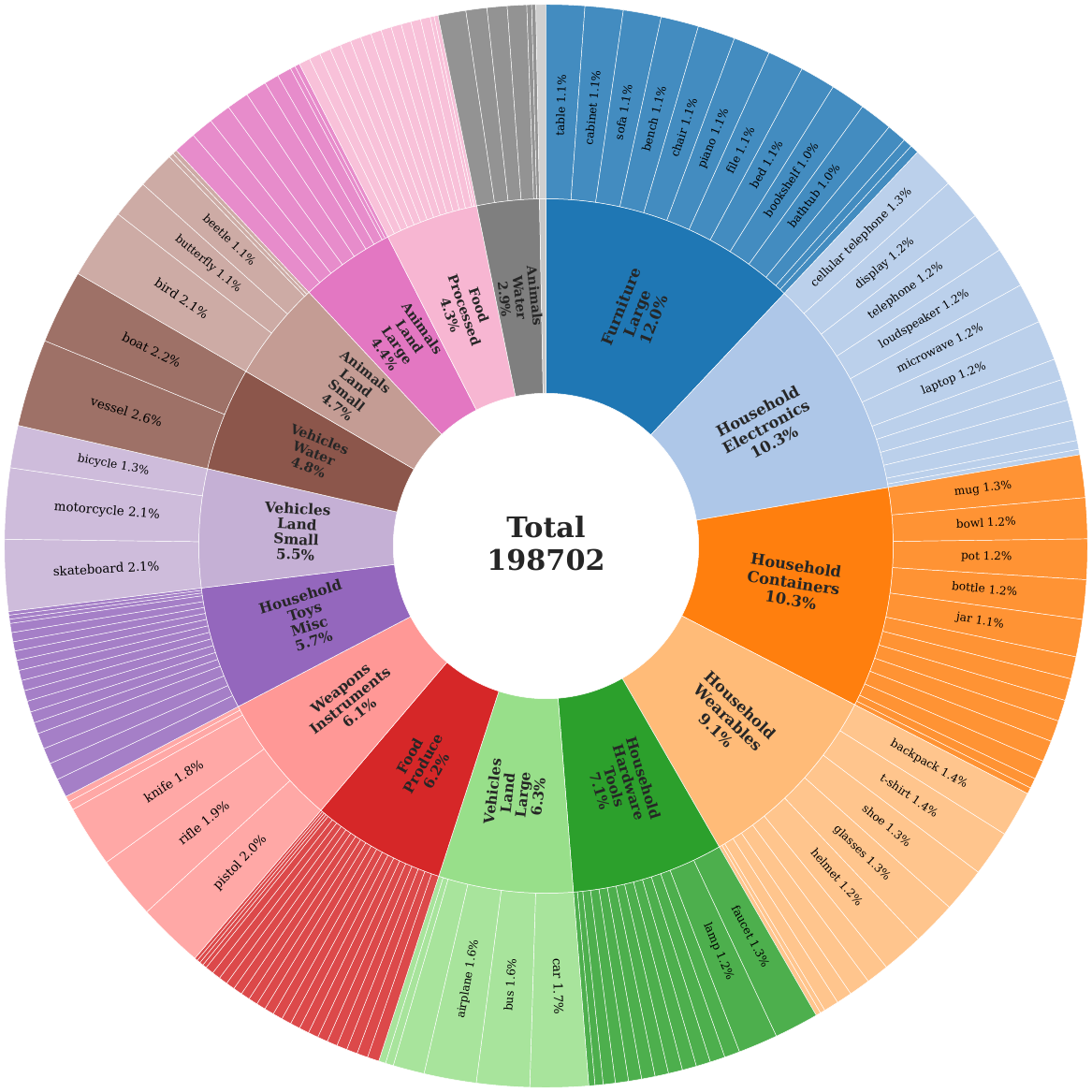}
        \caption{\textbf{Category distribution.}}
        \label{fig:kubricount_category_distribution}
    \end{subfigure}
    \hfill
    \begin{subfigure}[t]{0.59\textwidth}
        \centering
        \includegraphics[
            width=\linewidth,
            height=5.3cm,
            keepaspectratio
        ]{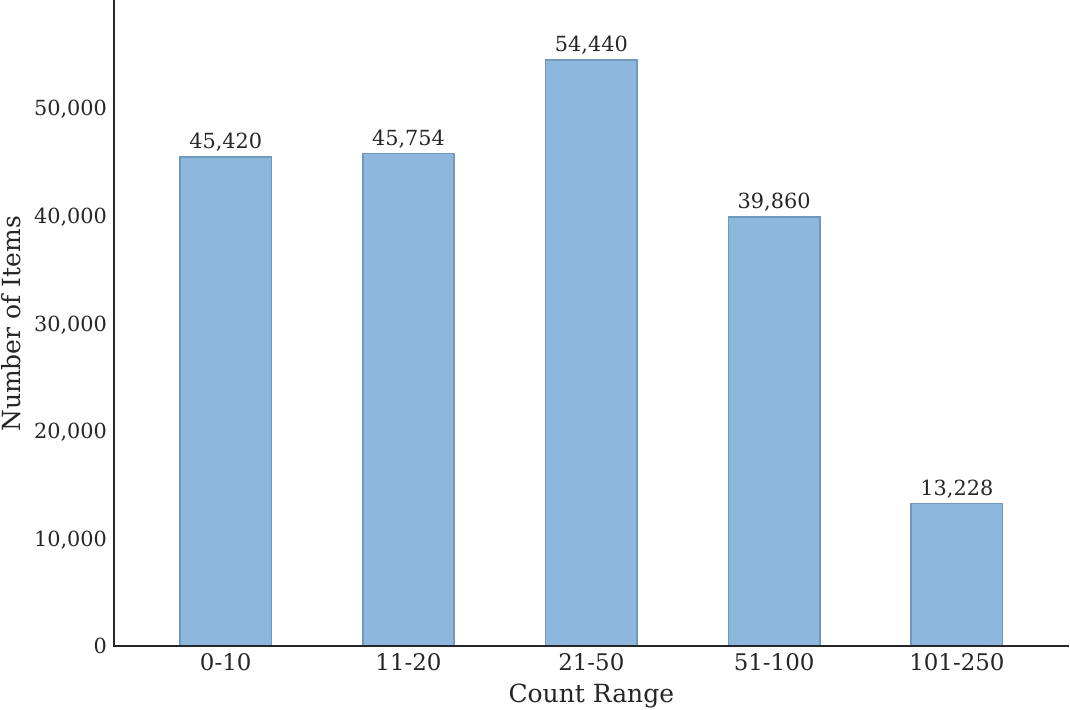}
        \caption{\textbf{Count distribution.} }
        \label{fig:kubricount_count_distribution}
    \end{subfigure}
    \vspace{-6pt}
    \caption{
        \textbf{KubriCount statistics.} 
        Category and count distributions show broad, balanced coverage for multi-grained evaluation. 
    }
    \label{fig:kubricount_statistics}
    \vspace{-14pt}
\end{figure*}

\vspace{3pt} 
\noindent \textbf{Data distributions.}
\cref{fig:kubricount_statistics} summarizes category and count statistics.
For Levels~2-5, each image yields two queries by swapping \(\mathcal{S}^+\) and \(\mathcal{S}^-\), producing around \textbf{198K} queries in total.
Counts range from sparse to crowded scenes and are capped at \textbf{250} instances per image (\cref{fig:kubricount_count_distribution}).
The category distribution is broad and well-balanced at both the super-category and sub-category levels~(\cref{fig:kubricount_category_distribution}), supporting rigorous multi-grained evaluation.

\subsection{Discussion}
\label{sec:dataset_discussion}
KubriCount is designed to evaluate counting under explicit control. 
We highlight five properties as follows.

\vspace{2pt}
\noindent \textbf{Scalable without manual annotations.}
KubriCount is built by a fully automatic pipeline and requires no human instance annotation, enabling dense, multi-object scenes at scale while retaining exact instance-level supervision.

\vspace{2pt}
\noindent \textbf{Controllable yet realistic.}
Synthesis gives direct control over categories, assets, and counts, producing a substantially more balanced benchmark than web-scraped data (\cref{fig:kubricount_statistics}).
Instead of forcing uniform counts, we try to preserve realistic frequency biases while avoiding extreme long-tail collapse.

\vspace{2pt}
\noindent \textbf{Multi-grained prompt-following.}
KubriCount encodes a five-level semantic hierarchy with controlled distractors that differ by exactly one factor ({\em e.g.}, attribute vs.\ category). This enables tests of whether a model counts the \emph{intended} set at the specified granularity, rather than exploiting single-category shortcuts.

\vspace{2pt}
\noindent \textbf{Rich supervision for analysis.}
Beyond scalar counts, each instance includes center points, 2D/3D boxes, and pixel-accurate masks (\cref{tab:dataset_comparison}), supporting diverse model designs and localization-aware error analysis.

\vspace{2pt}
\noindent \textbf{Leakage-free evaluation.}
Because all images are synthesized from 3D assets, KubriCount avoids overlap with common web-scale training corpora, providing a clean benchmark for evaluating foundation models.

%% file: tables/dataset_comparison.tex
\begin{table*}[t]
    \caption{
        \textbf{Comparison of counting datasets.}
        \textbf{\#Img} denotes the number of images, \textbf{\#Cat} denotes the number of categories, \textbf{\#Obj} denotes the total number of annotated objects, and \textbf{\#Max Obj} denotes the maximum number of objects in a single image.
    }
    \label{tab:dataset_comparison}
    \vspace{-4pt}
    \centering
    \scriptsize
    \setlength{\tabcolsep}{3pt}
    \renewcommand{\arraystretch}{1.2}
    \resizebox{\textwidth}{!}{
        \begin{tabular}{rccccccccc}
        \toprule
        \multirow{2}{*}{\textbf{Dataset}} & \multirow{2}{*}{\textbf{\#Img}} & \multirow{2}{*}{\textbf{\#Cat}} & \multirow{2}{*}{\textbf{\#Obj}} & \multirow{2}{*}{\textbf{\#Max Obj}} &
        \multicolumn{5}{c}{\textbf{Annotation}} \\
        \cline{6-10}
        &  &  &  &  & \textbf{Count} & \textbf{Point} & \textbf{Box} & \textbf{Mask} & \textbf{3D} \\
        \midrule
        UCF CC 50~\cite{ucfcc50}      & 50     & 1   & 64K  & 4543  & \textcolor{green}{\Checkmark} & \textcolor{green}{\Checkmark} & \textcolor{red}{\XSolidBrush} & \textcolor{red}{\XSolidBrush} & \textcolor{red}{\XSolidBrush} \\
        ShanghaiTech~\cite{shanghaitech}   & 1198   & 1   & 330K & 3139  & \textcolor{green}{\Checkmark} & \textcolor{green}{\Checkmark} & \textcolor{red}{\XSolidBrush} & \textcolor{red}{\XSolidBrush} & \textcolor{red}{\XSolidBrush} \\
        UCF QNRF~\cite{ucfqnrf}       & 1535   & 1   & 1.2M & 12865 & \textcolor{green}{\Checkmark} & \textcolor{green}{\Checkmark} & \textcolor{red}{\XSolidBrush} & \textcolor{red}{\XSolidBrush} & \textcolor{red}{\XSolidBrush} \\
        NWPU~\cite{nwpu}           & 5109   & 1   & 2.1M & 20033 & \textcolor{green}{\Checkmark} & \textcolor{green}{\Checkmark} & \textcolor{green}{\Checkmark} & \textcolor{red}{\XSolidBrush} & \textcolor{red}{\XSolidBrush} \\
        JHU Crowd~\cite{jhucrowd}      & 4372   & 1   & 1.5M & 25791 & \textcolor{green}{\Checkmark} & \textcolor{green}{\Checkmark} & \textcolor{green}{\Checkmark} & \textcolor{red}{\XSolidBrush} & \textcolor{red}{\XSolidBrush} \\
        CARPK~\cite{carpk}          & 1148   & 1   & 90K  & 188   & \textcolor{green}{\Checkmark} & \textcolor{green}{\Checkmark} & \textcolor{green}{\Checkmark} & \textcolor{red}{\XSolidBrush} & \textcolor{red}{\XSolidBrush} \\
        FSC-147~\cite{fsc147}        & 6135   & 147 & 344K & 3701  & \textcolor{green}{\Checkmark} & \textcolor{green}{\Checkmark} & \textcolor{red}{\XSolidBrush} & \textcolor{red}{\XSolidBrush} & \textcolor{red}{\XSolidBrush} \\
        REC-8K~\cite{groundingrec}         & 8011   & N/A & 287K & 1028  & \textcolor{green}{\Checkmark} & \textcolor{green}{\Checkmark} & \textcolor{red}{\XSolidBrush} & \textcolor{red}{\XSolidBrush} & \textcolor{red}{\XSolidBrush} \\
        OmniCount-191~\cite{omnicount}  & 30230  & 191 & 302K & 160   & \textcolor{green}{\Checkmark} & \textcolor{green}{\Checkmark} & \textcolor{green}{\Checkmark} & \textcolor{red}{\XSolidBrush} & \textcolor{red}{\XSolidBrush} \\
        TallyQA~\cite{tallyqa}        & 165443 & N/A & N/A  & 15    & \textcolor{green}{\Checkmark} & \textcolor{red}{\XSolidBrush} & \textcolor{red}{\XSolidBrush} & \textcolor{red}{\XSolidBrush} & \textcolor{red}{\XSolidBrush} \\
        CountBench~\cite{countbench}     & 540    & N/A & N/A  & 10    & \textcolor{green}{\Checkmark} & \textcolor{red}{\XSolidBrush} & \textcolor{red}{\XSolidBrush} & \textcolor{red}{\XSolidBrush} & \textcolor{red}{\XSolidBrush} \\
        Pixmo-Count~\cite{deitke2025molmo}    & 37996  & 365 & 139K & 10    & \textcolor{green}{\Checkmark} & \textcolor{green}{\Checkmark} & \textcolor{red}{\XSolidBrush} & \textcolor{red}{\XSolidBrush} & \textcolor{red}{\XSolidBrush} \\
        \midrule
        \rowcolor{cyan!10}
        \textbf{KubriCount~(Ours)} & \textbf{110507} & \textbf{157} & \textbf{7.3M} & \textbf{250} &
        \textbf{\textcolor{green}{\Checkmark}} & \textbf{\textcolor{green}{\Checkmark}} & \textbf{\textcolor{green}{\Checkmark}} & \textbf{\textcolor{green}{\Checkmark}} & \textbf{\textcolor{green}{\Checkmark}} \\
        \bottomrule
        \end{tabular}
    }
    \vspace{-6pt}
\end{table*}

%% file: sec/5_experiments.tex
\section{Experiments}

\subsection{Evaluation Settings}
\noindent\textbf{Benchmarks.}
We evaluate different counting scenarios on three benchmarks.
(i) \textbf{KubriCount} is our multi-grained counting benchmark, designed to test prompt following under explicit granularity (Levels~1--5) with controlled distractors; we report results on the union of TestA (unseen assets) and TestB (unseen categories) unless stated otherwise.
(ii) \textbf{FSC-147}~\cite{fsc147} evaluates class-agnostic exemplar-based counting in natural images.
(iii) \textbf{PairTally}~\cite{pairtally} evaluates prompt following in real-world multi-category scenes with hard negatives.

\vspace{3pt}
\noindent\textbf{Baselines.}
On KubriCount, we benchmark both (i) multimodal large language models (MLLMs) via direct prompting and (ii) counting expert models with explicit localization.
MLLMs include open-source families~\cite{Qwen2.5-VL,Qwen3-VL,InternVL2.5,InternVL3,InternVL3.5,deitke2025molmo,clark2026molmo2,llava,llavaonevision,llama3,llama-cot}, 
{\em e.g.}, Qwen-VL, InternVL, Molmo/Molmo2, LLaVA, and LLaMA-3.2V, 
and proprietary APIs~\cite{gemini2.5,gemini3,gpt4o,gpt5,GPT-5.1,GPT-5.2,haiku45,sonnet45,opus45}, 
{\em e.g.}, Gemini, GPT, and Claude.
Expert models include density-regression methods~\cite{fsc147,countr,loca} (FamNet, CounTR, LoCA), detection-based methods~\cite{pelhan2024dave,pelhan2024geco,countgd,countgd++} (DAVE, GeCo, CountGD, CountGD++), and Rex-omni~\cite{rexomni}.
On FSC-147 and PairTally, we focus on comparing counting expert models and our HieraCount, as these benchmarks are primarily defined around exemplar-based protocols.

\vspace{3pt}
\noindent\textbf{Evaluation metrics.}
Following common practice, we report Mean Absolute Error (\textbf{MAE}) and Root Mean Squared Error (\textbf{RMSE}) between predicted and ground-truth counts on all benchmarks, with lower values indicating better accuracy.
Notably, given KubriCount's point/box/mask annotations, future work may adopt localization-aware metrics for further analysis.

\vspace{3pt}
\noindent\textbf{Evaluation protocols.}
On KubriCount, MLLMs are prompted to output a single integer (no localization).
We use text prompts for Levels~1/2/3/5 (including negative text when distractors exist); for Level~4, we additionally provide exemplar bounding boxes as text coordinates to indicate the target instance type.
Counting expert models (and HieraCount) are evaluated under their native prompting interfaces, which are primarily \emph{exemplar-based}; only the CountGD family additionally supports text prompts.
On FSC-147, we follow the standard protocol using three visual exemplars and a text prompt.
On PairTally, we report results under \emph{positive-only} prompting for fair comparison, even though negative prompts can further improve performance.

\input{tables/kubricount_results}

\subsection{Comparison to State-of-the-Art}
\noindent\textbf{KubriCount benchmark results.}
Tab.~\ref{tab:kubricount_results} summarizes multi-grained results on KubriCount for representative MLLMs, expert models, and our HieraCount.

For MLLMs, we observe a consistent gap between open-source and proprietary systems: commercial models outperform open-source MLLMs overall, with Gemini-3-Flash/Pro~\cite{gemini3} achieving the best MAE, while the best open-source models (InternVL3-78B~\cite{InternVL3} and LLaMA-3.2V-11B~\cite{llama3}) remain notably behind.
Within each open-source family, performance generally improves with model scale, although strong mid-scale models can already be competitive.
Across levels, errors on Levels~2/3/5 are comparable, while Level~1 is often more challenging due to larger target counts.
Level~4 is the primary failure mode: fine-grained within-category discrimination and box grounding can trigger occasional catastrophic errors (e.g., MolmoE-1B~\cite{deitke2025molmo} and LLaVA-1.5~\cite{llava}).
CoT variants do not consistently improve performance~({\em e.g.}, LLaMA-3.2V-CoT~\cite{llama-cot}).

Expert models are substantially stronger than MLLMs on Level~1 and can approach top proprietary performance.
However, most methods degrade markedly on Levels~2--5, indicating limited prompt following in the presence of distractors and fine-grained distinctions.
CountGD++ underscores the value of explicit exclusion: negative prompts substantially improve performance on the harder levels, suggesting that much of today’s controllability comes from rejecting distractors rather than robust positive matching.
We also observe occasional large-error outliers for detection-based methods, pointing to remaining robustness issues in dense scenes.
Trained with granularity-aware prompts, HieraCount achieves the best performance under \emph{positive-only} prompting and improves substantially over prior expert models across levels.
Level~4 remains the most challenging for HieraCount due to fine-grained within-category instance-type discrimination.

\input{tables/fsc147_pairtally}

\vspace{3pt}
\noindent\textbf{FSC-147 results.}
We evaluate HieraCount on FSC-147~\cite{fsc147} under the standard protocol.
As shown in~\cref{tab:fsc147_pairtally}, HieraCount achieves mid-range MAE and remains reasonably robust, but does not outperform the strongest FSC-147-specialized baselines. We attribute the gap to (i) a systematic mismatch in task protocol: FSC-147 largely reflects identity-level counting without explicit distractors, whereas our training targets stricter matching under multi-grained semantics with controlled distractors, and (ii) inference robustness of detection-based counting in dense scenes, which disproportionately affects RMSE on several extremely high-count images.

\vspace{3pt}
\noindent\textbf{Generalization to PairTally.}
We further evaluate HieraCount on PairTally~\cite{pairtally}, a challenging real-world benchmark with multi-category scenes and hard negatives.
Unlike FSC-147, PairTally aligns closely with our multi-grained protocol, allowing HieraCount to better showcase generalization.
As shown in~\cref{tab:fsc147_pairtally}, HieraCount substantially improves over CountGD and achieves state-of-the-art performance under \emph{positive-only} prompting. These results suggest that our pipeline effectively mitigates the sim-to-real gap, and that multi-grained supervision with controlled distractors improves prompt following in cluttered real-world scenes.

\begin{figure*}[t]
  \centering
  \includegraphics[width=\textwidth]{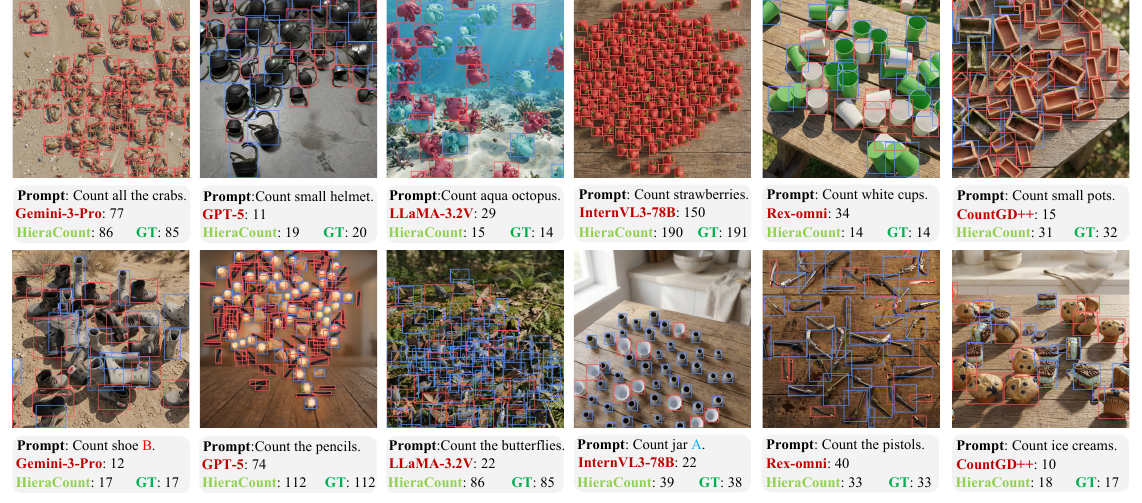}
  \vspace{-16pt}
  \caption{
    \textbf{Qualitative analysis.}
    Representative failure cases of prior models under multi-grained prompts, contrasted with HieraCount under the same prompts.
  }
  \label{fig:qualitative}
  \vspace{-16pt}
\end{figure*}

\vspace{3pt}
\noindent \textbf{Qualitative analysis.}
As shown in~\cref{fig:qualitative}, we visualize representative failure cases of prior models and contrast them with HieraCount’s improvements under the same prompts.
Each panel shows the prompt, a baseline model prediction, HieraCount's prediction, and the ground-truth count (GT), highlighting improved prompt following on challenging attribute-/instance-sensitive queries.

%% file: tables/kubricount_results.tex
\begin{table*}[!t]
    \caption{
        \textbf{KubriCount Benchmark Results.}
        We report MAE/RMSE for each level and overall, with the best and second-best \setlength{\fboxsep}{2pt}\colorbox{mygreen}{\textbf{bolded}} and \setlength{\fboxsep}{1.5pt}\colorbox{mylightgreen}{\underline{underlined}}, respectively.
    }
    \label{tab:kubricount_results}
    \vspace{-4pt}

    \centering
    \scriptsize
    \setlength{\tabcolsep}{1.9pt}
    \renewcommand{\arraystretch}{0.95}
    \resizebox{\textwidth}{!}{
    \begin{tabular}{lccc*{5}{cc}}
    \toprule[1.5pt]
    \textbf{Methods} & \multicolumn{2}{c}{\textbf{Overall}} & \textbf{Rank}
    & \multicolumn{2}{c}{\textbf{L1}} & \multicolumn{2}{c}{\textbf{L2}} & \multicolumn{2}{c}{\textbf{L3}} & \multicolumn{2}{c}{\textbf{L4}} & \multicolumn{2}{c}{\textbf{L5}} \\
    \cmidrule(lr){2-3}\cmidrule(lr){5-6}\cmidrule(lr){7-8}\cmidrule(lr){9-10}\cmidrule(lr){11-12}\cmidrule(lr){13-14}
    & \textbf{MAE} & \textbf{RMSE} & &
    \textbf{MAE} & \textbf{RMSE} &
    \textbf{MAE} & \textbf{RMSE} &
    \textbf{MAE} & \textbf{RMSE} &
    \textbf{MAE} & \textbf{RMSE} &
    \textbf{MAE} & \textbf{RMSE} \\
    \midrule[0.8pt]

    \rowcolor{cyan!10}
    \multicolumn{14}{c}{\textbf{\textit{Small-scale MLLMs (1B$\sim$4B)}}} \\
    \midrule
    MolmoE-1B-0924~\cite{deitke2025molmo}           & 40.36 & 156.13 & 11 & 19.50 & 32.77 & \cellcolor{mylightgreen}\underline{11.36} & \cellcolor{mylightgreen}\underline{19.60} & 11.18 & 19.30 & 148.80 & 346.84 & \cellcolor{mylightgreen}\underline{11.71} & 19.90 \\
    InternVL2.5-1B~\cite{InternVL2.5}            & 20.02 & 30.99 & 10 & 28.32 & 43.56 & 18.50 & 27.67 & 16.81 & 25.97 & 19.41 & 28.43 & 16.92 & 25.48 \\
    InternVL3-1B~\cite{InternVL3}             & 18.55 & 29.27 & 9 & 24.32 & 39.62 & 17.76 & 26.58 & 15.39 & 24.67 & 19.98 & 28.60 & 15.10 & 23.91 \\
    InternVL3.5-1B~\cite{InternVL3.5}           & 14.48 & 24.41 & 5  & 18.73 & 32.44 & 11.79 & 19.96 & 12.30 & 20.97 & 17.18 & 25.80 & 12.46 & 20.70 \\
    Qwen3-VL-2B~\cite{Qwen3-VL}              & 16.43 & 25.38 & 7  & 16.09 & 26.73 & 16.02 & 24.65 & 14.58 & 23.14 & 20.31 & 28.79 & 15.05 & 23.02 \\
    SpaceQwen-3B~\cite{chen2024spatialvlm}             & 16.61 & 25.97 & 8  & 22.82 & 33.45 & 14.60 & 23.44 & 14.66 & 23.48 & 16.25 & 24.82 & 14.73 & 23.17 \\
    Qwen2.5-VL-3B~\cite{Qwen2.5-VL}            & 15.77 & 25.97 & 6  & 15.64 & 26.04 & 18.33 & 29.51 & 14.99 & 25.11 & 15.63 & 26.11 & 13.97 & 22.08 \\
    Qwen3-VL-4B~\cite{Qwen3-VL}              & \cellcolor{mygreen}\textbf{12.22} & \cellcolor{mygreen}\textbf{21.15} & \cellcolor{mygreen}\textbf{1}  & \cellcolor{mygreen}\textbf{12.50} & \cellcolor{mylightgreen}\underline{23.62} & 12.46 & 21.62 & 10.98 & \cellcolor{mylightgreen}\underline{18.70} & 13.04 & \cellcolor{mylightgreen}\underline{22.08} & 12.06 & \cellcolor{mylightgreen}\underline{19.12} \\
    InternVL2.5-4B~\cite{InternVL2.5}           & 13.23 & \cellcolor{mylightgreen}\underline{21.81} & 4  & \cellcolor{mylightgreen}\underline{13.86} & \cellcolor{mygreen}\textbf{22.75} & 13.25 & 21.97 & 12.86 & 20.92 & \cellcolor{mygreen}\textbf{12.41} & \cellcolor{mygreen}\textbf{22.02} & 13.75 & 21.28 \\
    InternVL3.5-4B~\cite{InternVL3.5}            & 13.22 & 22.07 & 3  & 14.47 & 25.73 & 12.12 & 20.23 & \cellcolor{mylightgreen}\underline{10.37} & \cellcolor{mygreen}\textbf{17.84} & 16.34 & 25.50 & 12.77 & 19.78 \\
    Molmo2-4B~\cite{clark2026molmo2}                & \cellcolor{mylightgreen}\underline{12.99} & 26.06 & \cellcolor{mylightgreen}\underline{2}  & 24.43 & 44.36 & \cellcolor{mygreen}\textbf{8.09} & \cellcolor{mygreen}\textbf{16.23} & \cellcolor{mygreen}\textbf{9.18} & 18.83 & \cellcolor{mylightgreen}\underline{12.91} & 22.16 & \cellcolor{mygreen}\textbf{10.51} & \cellcolor{mygreen}\textbf{18.05} \\
    \midrule[0.8pt]

    \rowcolor{cyan!10}
    \multicolumn{14}{c}{\textbf{\textit{Mid-scale MLLMs (7B$\sim$14B)}}} \\
    \midrule
    LLaVA-1.5-7B~\cite{llava}             & 28.13 & 91.98 & 18 & 23.16 & 34.10 & 15.42 & 24.71 & 15.59 & 25.57 & 71.48 & 198.72 & 15.32 & 24.30 \\
    LLaVA-OV-7B~\cite{llavaonevision}              & 14.05 & 23.25 & 16 & 15.76 & 27.37 & 13.74 & 22.04 & 13.64 & 22.32 & 12.55 & 21.42 & 14.59 & 22.65 \\
    Qwen2.5-VL-7B~\cite{Qwen2.5-VL}            & 11.52 & 20.15 & 5  & 12.07 & 23.40 & 10.89 & 19.07 & 9.90 & 17.85 & 13.31 & 21.77 & 11.39 & 17.99 \\
    SpaceR-7B~\cite{ouyang2025spacer}                & 12.29 & 23.11 & 12 & 13.03 & 24.10 & 11.38 & 19.16 & 10.98 & 18.53 & 13.48 & 31.60 & 12.62 & 19.60 \\
    Molmo-7B-o-0924~\cite{deitke2025molmo}          & 13.50 & 24.83 & 14 & 17.58 & 33.82 & 11.64 & 20.88 & 11.01 & 20.54 & 15.16 & 25.18 & 12.09 & 21.16 \\
    Molmo-7B-d-0924~\cite{deitke2025molmo}          & 12.13 & 21.58 & 11 & 13.83 & 26.04 & 11.77 & 20.72 & 10.24 & 18.69 & 13.44 & 22.63 & 11.30 & 18.78 \\
    Molmo2-o-7B ~\cite{clark2026molmo2}             & 11.66 & 20.62 & 7  & 13.96 & 26.05 & 10.41 & 18.46 & 9.15 & 16.52 & 13.20 & 21.91 & 11.58 & 18.73 \\
    InternVL2.5-8B~\cite{InternVL2.5}           & 13.15 & 21.22 & 13 & 14.05 & 22.89 & 12.01 & 19.41 & 11.67 & 18.97 & 14.42 & 23.57 & 13.64 & 20.90 \\
    InternVL3-8B~\cite{InternVL3}              & 11.97 & 19.92 & 9  & 11.89 & 21.18 & 11.92 & 19.96 & 10.86 & 18.13 & 12.66 & 20.83 & 12.50 & 19.26 \\
    InternVL3.5-8B~\cite{InternVL3.5}            & 11.53 & 19.73 & 6  & 12.00 & 22.37 & 11.25 & 18.96 & 9.63 & 16.70 & 12.70 & 21.37 & 12.02 & 18.63 \\
    Qwen3-VL-8B~\cite{Qwen3-VL}              & 11.99 & 20.94 & 10 & 10.05 & 19.70 & 13.07 & 22.32 & 12.04 & 20.99 & 12.66 & 21.92 & 12.06 & 19.48 \\
    Molmo2-8B~\cite{clark2026molmo2}                 & 11.67 & 21.75 & 8  & 15.29 & 30.46 & 9.98 & 17.90 & 9.72 & 18.09 & 12.04 & 20.84 & 11.33 & 18.92 \\
    LLaMA-3.2V-11B~\cite{llama3}           & \cellcolor{mygreen}\textbf{8.85} & \cellcolor{mygreen}\textbf{15.55} & \cellcolor{mygreen}\textbf{1}  & \cellcolor{mygreen}\textbf{8.89} & \cellcolor{mylightgreen}\underline{17.89} & \cellcolor{mylightgreen}\underline{8.66} & \cellcolor{mygreen}\textbf{14.36} & \cellcolor{mylightgreen}\underline{8.08} & \cellcolor{mylightgreen}\underline{14.30} & \cellcolor{mygreen}\textbf{8.84} & \cellcolor{mygreen}\textbf{14.77} & \cellcolor{mygreen}\textbf{9.78} & \cellcolor{mylightgreen}\underline{16.18} \\
    LLaMA-3.2V-11B-CoT~\cite{llama-cot}       & 11.27 & 19.31 & 4  & \cellcolor{mylightgreen}\underline{9.62} & 20.42 & 10.27 & 17.00 & 8.78 & 15.40 & 16.31 & 24.56 & 11.35 & 17.85 \\
    LLaVA-1.5-13B~\cite{llava}            & 16.12 & 28.94 & 17 & 18.49 & 28.39 & 14.49 & 23.90 & 15.14 & 25.03 & 18.13 & 40.56 & 14.40 & 23.40 \\
    InternVL3-14B~\cite{InternVL3}            & 10.52 & 17.74 & 3  & 11.81 & 20.12 & 9.16 & 15.97 & 9.16 & 15.55 & 11.07 & 18.89 & 11.48 & 17.78 \\
    InternVL3.5-14B~\cite{InternVL3.5}          & \cellcolor{mylightgreen}\underline{9.08} & \cellcolor{mylightgreen}\underline{15.58} & \cellcolor{mylightgreen}\underline{2}  & 9.96 & \cellcolor{mygreen}\textbf{17.40} & \cellcolor{mygreen}\textbf{8.20} & \cellcolor{mylightgreen}\underline{15.12} & \cellcolor{mygreen}\textbf{7.50} & \cellcolor{mygreen}\textbf{13.76} & \cellcolor{mylightgreen}\underline{9.45} & \cellcolor{mylightgreen}\underline{15.32} & \cellcolor{mylightgreen}\underline{10.33} & \cellcolor{mygreen}\textbf{16.02} \\
    Kimi-VL-16B-A3B~\cite{kimivl}          & 13.64 & 22.20 & 15  & 13.55 & 23.96 & 13.61 & 21.92 & 12.76 & 20.20 & 13.82 & 22.88 & 14.43 & 21.77 \\
    \midrule[0.8pt]

    \rowcolor{cyan!10}
    \multicolumn{14}{c}{\textbf{\textit{Large-scale MLLMs (30B$\sim$78B and beyond)}}} \\
    \midrule
    Qwen3-VL-30B-A3B~\cite{Qwen3-VL}          & 13.24 & 23.23 & 11 & 13.01 & 28.70 & 13.13 & 22.05 & 11.75 & 19.66 & 15.16 & 23.88 & 13.11 & 20.61 \\
    Qwen2.5-VL-32B~\cite{Qwen2.5-VL}           & 9.07 & 16.55 & 3  & 9.60 & 17.87 & 9.45 & 18.34 & 8.21 & 15.50 & \cellcolor{mylightgreen}\underline{8.56} & \cellcolor{mylightgreen}\underline{15.43} & \cellcolor{mygreen}\textbf{9.48} & \cellcolor{mygreen}\textbf{15.08} \\
    Qwen3-VL-32B~\cite{Qwen3-VL}              & 11.23 & 20.18 & 9  & 12.05 & 22.96 & 10.06 & 18.67 & 11.00 & 19.97 & 11.37 & 19.99 & 11.80 & 19.09 \\
    InternVL2.5-38B~\cite{InternVL2.5}          & 11.31 & 18.99 & 10 & 13.60 & 22.63 & 9.14 & 15.90 & 9.49 & 15.85 & 12.40 & 21.14 & 12.05 & 18.56 \\
    InternVL3-38B~\cite{InternVL3}            & 9.57 & 16.55 & 5  & 10.91 & 19.87 & 6.69 & \cellcolor{mygreen}\textbf{12.39} & 8.15 & \cellcolor{mylightgreen}\underline{13.84} & 11.33 & 18.72 & 10.97 & 16.92 \\
    InternVL3.5-38B~\cite{InternVL3.5}           & 10.24 & 18.43 & 7  & 9.68 & 19.75 & 9.29 & 16.64 & 8.32 & 15.27 & 13.27 & 22.51 & 10.66 & 17.06 \\
    LLaVA-OV-72B~\cite{llavaonevision}             & 11.22 & 19.45 & 8  & 12.00 & 21.74 & 8.74 & 15.87 & 9.65 & 16.82 & 14.04 & 23.17 & 11.83 & 18.82 \\
    Qwen2.5-VL-72B~\cite{Qwen2.5-VL}           & \cellcolor{mylightgreen}\underline{8.95} & 21.89 & \cellcolor{mylightgreen}\underline{2}  & 9.86 & 28.97 & 8.89 & 27.77 & 8.26 & 15.98 & \cellcolor{mygreen}\textbf{7.99} & \cellcolor{mygreen}\textbf{14.92} & \cellcolor{mylightgreen}\underline{9.75} & 16.28 \\
    InternVL2.5-78B~\cite{InternVL2.5}          & 9.17 & \cellcolor{mylightgreen}\underline{16.09} & 4  & \cellcolor{mylightgreen}\underline{8.97} & \cellcolor{mylightgreen}\underline{16.35} & \cellcolor{mylightgreen}\underline{7.72} & 14.06 & \cellcolor{mygreen}\textbf{7.53} & \cellcolor{mygreen}\textbf{13.30} & 11.38 & 19.51 & 10.35 & 16.54 \\
    InternVL3-78B~\cite{InternVL3}            & \cellcolor{mygreen}\textbf{8.82} & \cellcolor{mygreen}\textbf{15.67} & \cellcolor{mygreen}\textbf{1}  & \cellcolor{mygreen}\textbf{8.42} & \cellcolor{mygreen}\textbf{15.91} & \cellcolor{mygreen}\textbf{7.24} & \cellcolor{mylightgreen}\underline{13.57} & \cellcolor{mylightgreen}\underline{7.67} & 13.87 & 10.94 & 18.47 & 9.96 & \cellcolor{mylightgreen}\underline{16.15} \\
    Qwen3-VL-235B-A22B~\cite{Qwen3-VL}        & 9.95 & 18.47 & 6  & 9.71 & 20.92 & 8.80 & 16.70 & 9.60 & 18.25 & 10.69 & 18.10 & 11.06 & 18.24 \\
    \midrule[0.8pt]

    \rowcolor{cyan!10}
    \multicolumn{14}{c}{\textbf{\textit{Proprietary MLLMs (Commercial APIs)}}} \\
    \midrule
    Gemini-2.5-Pro~\cite{gemini2.5}           & 10.71 & 25.37 & 8  & 15.96 & 43.42 & 11.42 & 22.28 & 8.14 & 15.69 & 10.23 & 19.82 & 7.50 & \cellcolor{mygreen}\textbf{13.26} \\
    Gemini-3-Flash~\cite{gemini3}           & \cellcolor{mygreen}\textbf{5.41} & 26.35 & \cellcolor{mygreen}\textbf{1}  & \cellcolor{mygreen}\textbf{5.99} & 48.91 & \cellcolor{mylightgreen}\underline{4.59} & 18.17 & \cellcolor{mygreen}\textbf{4.23} & 17.82 & \cellcolor{mygreen}\textbf{6.32} & \cellcolor{mygreen}\textbf{14.01} & \cellcolor{mygreen}\textbf{5.97} & 15.16 \\
    Gemini-3-Pro~\cite{gemini3}             & \cellcolor{mylightgreen}\underline{5.49} & 27.05 & \cellcolor{mylightgreen}\underline{2}  & \cellcolor{mylightgreen}\underline{6.10} & 51.13 & \cellcolor{mygreen}\textbf{4.32} & \cellcolor{mygreen}\textbf{11.30} & \cellcolor{mylightgreen}\underline{4.37} & 19.76 & \cellcolor{mylightgreen}\underline{6.49} & 16.81 & \cellcolor{mylightgreen}\underline{6.23} & 16.17 \\
    GPT-4o~\cite{gpt4o}                   & 9.40 & 16.67 & 6  & 10.41 & \cellcolor{mylightgreen}\underline{21.00} & 7.50 & 13.58 & 8.51 & \cellcolor{mylightgreen}\underline{14.73} & 10.43 & 16.81 & 10.37 & 16.39 \\
    GPT-5~\cite{gpt5}                    & 7.81 & \cellcolor{mygreen}\textbf{15.01} & 3  & 8.67 & \cellcolor{mygreen}\textbf{18.91} & 6.41 & 12.18 & 6.75 & \cellcolor{mygreen}\textbf{12.71} & 8.84 & 15.98 & 8.55 & \cellcolor{mylightgreen}\underline{14.40} \\
    GPT-5.1~\cite{GPT-5.1}                  & 8.30 & 16.57 & 4  & 9.36 & 21.80 & 6.94 & 13.99 & 7.60 & 14.99 & 8.78 & 16.07 & 8.93 & 14.87 \\
    GPT-5.2~\cite{GPT-5.2}                  & 9.44 & 25.99 & 7  & 11.99 & 21.12 & 6.62 & \cellcolor{mylightgreen}\underline{11.90} & 8.64 & 48.36 & 9.73 & 16.57 & 10.46 & 16.40 \\
    Claude-4.5-Haiku~\cite{haiku45}         & 11.47 & 24.32 & 9  & 14.39 & 28.48 & 8.92 & 16.67 & 10.23 & 18.32 & 11.61 & 34.05 & 12.36 & 19.59 \\
    Claude-4.5-Sonnet~\cite{sonnet45}        & 11.61 & 26.17 & 10 & 14.46 & 28.37 & 8.98 & 16.74 & 10.17 & 18.00 & 12.25 & 40.55 & 12.34 & 19.60 \\
    Claude-4.5-Opus~\cite{opus45}          & 8.87 & \cellcolor{mylightgreen}\underline{16.53} & 5  & 10.67 & 21.25 & 7.81 & 14.20 & 8.17 & 15.53 & 8.83 & \cellcolor{mylightgreen}\underline{15.66} & 8.93 & 15.17 \\
    \midrule[0.8pt]

    \rowcolor{cyan!10}
    \multicolumn{14}{c}{\textbf{\textit{Counting expert models}}} \\
    \midrule
    FamNet~\cite{fsc147}                    & 21.17 & 37.49 & 9 & 21.55 & 30.72 & 26.85 & 50.57 & 20.09 & 35.49 & 16.98 & 27.83 & 19.77 & 37.21 \\
    LoCA~\cite{loca}                        & 16.69 & 29.07 & 6 & 9.80 & 15.60 & 24.15 & 39.30 & 17.31 & 32.93 & 16.78 & 26.87 & 14.74 & 23.99 \\
    DAVE~\cite{pelhan2024dave}              & 18.04 & 31.51 & 7 & 12.18 & 19.68 & 22.22 & 39.34 & 19.71 & 39.18 & 17.95 & 26.94 & 17.86 & 27.20 \\
    GeCo~\cite{pelhan2024geco}              & 10.82 & \cellcolor{mylightgreen}\underline{18.63} & 3 & 8.23 & \cellcolor{mylightgreen}\underline{14.98} & 13.33 & \cellcolor{mylightgreen}\underline{22.97} & 10.19 & \cellcolor{mylightgreen}\underline{17.89} & 12.49 & \cellcolor{mylightgreen}\underline{19.97} & 9.58 & 15.68 \\
    CounTR~\cite{countr}                    & 12.82 & 21.96 & 4 & 8.86 & 15.77 & 15.42 & 27.86 & 13.50 & 22.32 & 14.07 & 22.13 & 12.07 & 19.30 \\
    CountGD~\cite{countgd}                  & 18.18 & 38.49 & 8 & 7.67 & 43.86 & 24.51 & 47.76 & 18.89 & 36.11 & 22.49 & 33.09 & 16.80 & 26.52 \\
    CountGD++~\cite{countgd++}              & \cellcolor{mylightgreen}\underline{7.76} & 28.17 & \cellcolor{mylightgreen}\underline{2} & 7.65 & 45.06 & \cellcolor{mylightgreen}\underline{8.74} & 26.05 & \cellcolor{mylightgreen}\underline{6.66} & 21.97 & \cellcolor{mylightgreen}\underline{9.01} & 24.02 & \cellcolor{mylightgreen}\underline{6.55} & \cellcolor{mylightgreen}\underline{13.09} \\
    Rex-omni~\cite{rexomni}                 & 14.66 & 42.78 & 5 & \cellcolor{mylightgreen}\underline{6.89} & 37.86 & 23.34 & 51.88 & 12.01 & 42.24 & 18.29 & 42.98 & 11.87 & 36.10 \\
    \textbf{HieraCount~(Ours)}                          & \cellcolor{mygreen}\textbf{4.67} & \cellcolor{mygreen}\textbf{11.07} & \cellcolor{mygreen}\textbf{1} & \cellcolor{mygreen}\textbf{3.06} & \cellcolor{mygreen}\textbf{10.58} & \cellcolor{mygreen}\textbf{3.10} & \cellcolor{mygreen}\textbf{7.66} & \cellcolor{mygreen}\textbf{3.90} & \cellcolor{mygreen}\textbf{8.29} & \cellcolor{mygreen}\textbf{8.37} & \cellcolor{mygreen}\textbf{17.14} & \cellcolor{mygreen}\textbf{5.04} & \cellcolor{mygreen}\textbf{9.08} \\
    \bottomrule[1.5pt]
    \end{tabular}
    }
    \vspace{-8pt}
\end{table*}

%% file: tables/fsc147_pairtally.tex
\begin{table*}[t]
    \centering
    \caption{\textbf{HieraCount generalization to FSC-147 and PairTally.} Text/Box denote text prompt and box exemplar.}
    \label{tab:fsc147_pairtally}
    \vspace{-4pt}
    \scriptsize
    \setlength{\tabcolsep}{3.8pt}
    \renewcommand{\arraystretch}{1.05}
    \resizebox{\textwidth}{!}{
    \begin{tabular}{@{}lcc|cc|cc|cc@{}}
    \toprule
    \textbf{Method} & \textbf{Text} & \textbf{Box}
    & \multicolumn{2}{c|}{\textbf{FSC-147 Val}}
    & \multicolumn{2}{c|}{\textbf{FSC-147 Test}}
    & \multicolumn{2}{c}{\textbf{PairTally}} \\
    \cmidrule(lr){4-5}\cmidrule(lr){6-7}\cmidrule(lr){8-9}
    & & & \textbf{MAE} & \textbf{RMSE} & \textbf{MAE} & \textbf{RMSE} & \textbf{MAE} & \textbf{RMSE} \\
    \midrule
    FamNet~\cite{fsc147}              & \textcolor{red}{\XSolidBrush} & \textcolor{green}{\Checkmark} & 23.75 & 69.07 & 22.08 & 99.54 & 75.83 & -- \\
    LoCA~\cite{loca}                  & \textcolor{red}{\XSolidBrush} & \textcolor{green}{\Checkmark} & 10.24 & 32.56 & 10.79 & 56.97 & 72.80 & -- \\
    GeCo~\cite{pelhan2024geco}        & \textcolor{red}{\XSolidBrush} & \textcolor{green}{\Checkmark} & 9.52 & 43.00 & 7.91 & 54.28 & 50.24 & -- \\
    DAVE~\cite{pelhan2024dave}        & \textcolor{red}{\XSolidBrush} & \textcolor{green}{\Checkmark} & 8.91 & 28.08 & 8.66 & 32.36 & 47.37 & -- \\
    CountGD~\cite{countgd}            & \textcolor{green}{\Checkmark} & \textcolor{green}{\Checkmark} & \textbf{7.10} & \textbf{26.08} & \textbf{5.74} & \textbf{24.09} & 46.67 & 70.85 \\
    CountGD++~\cite{countgd++}        & \textcolor{green}{\Checkmark} & \textcolor{green}{\Checkmark} & -- & -- & 7.95 & 29.24 & 46.41 & 69.52 \\
    \midrule
    \textbf{Our HieraCount}           & \textcolor{green}{\Checkmark} & \textcolor{green}{\Checkmark} & 10.83 & 49.36 & 10.29 & 99.13 & \textbf{36.27} & \textbf{65.76} \\
    \bottomrule
    \end{tabular}
    }
    \vspace{-6pt}
\end{table*}

%% file: sec/6_conclusion.tex
\section{Conclusion}
We take a data-centric step toward robust and controllable open-world counting.
Motivated by the gap between existing category-level counting formulations and the multi-level semantic hierarchy of real-world objects, we define a \textbf{multi-grained counting} task that makes granularity explicit and verifiable.
To overcome the data bottleneck, we propose the first fully automatic pipeline for scaling counting data and build \textbf{KubriCount}, to our knowledge the largest and most comprehensively annotated counting benchmark to date.
Extensive evaluations of MLLMs and expert models reveal persistent limitations under fine-grained distinctions.
Finally, we introduce \textbf{HieraCount}, trained with granularity-aware prompts on KubriCount, and show substantial gains together with strong real-world generalization.
We hope this formulation, pipeline, benchmark, and model will support future work on scalable and reliable multi-grained counting.


%% file: sec/X_supp.tex
\clearpage

\begin{center}
    {\LARGE \bfseries Count Anything at Any Granularity\\[10pt]}
    {\large Supplementary Material}
\end{center}

\setcounter{tocdepth}{2} 

\doublespacing
\tableofcontentsonthispage
\singlespacing

\clearpage

\section{Qualitative Visualizations}

We provide qualitative visualizations from \textbf{KubriCount} across all five levels. For each level, we show a diverse set of examples to demonstrate the scene complexity, category coverage, target-distractor design, and annotation quality of the dataset. These examples complement the visualizations in the main paper and provide a clear view of how different semantic granularity levels are instantiated in our benchmark.

\begin{figure*}[!htb]
    \centering
    \includegraphics[width=\textwidth]{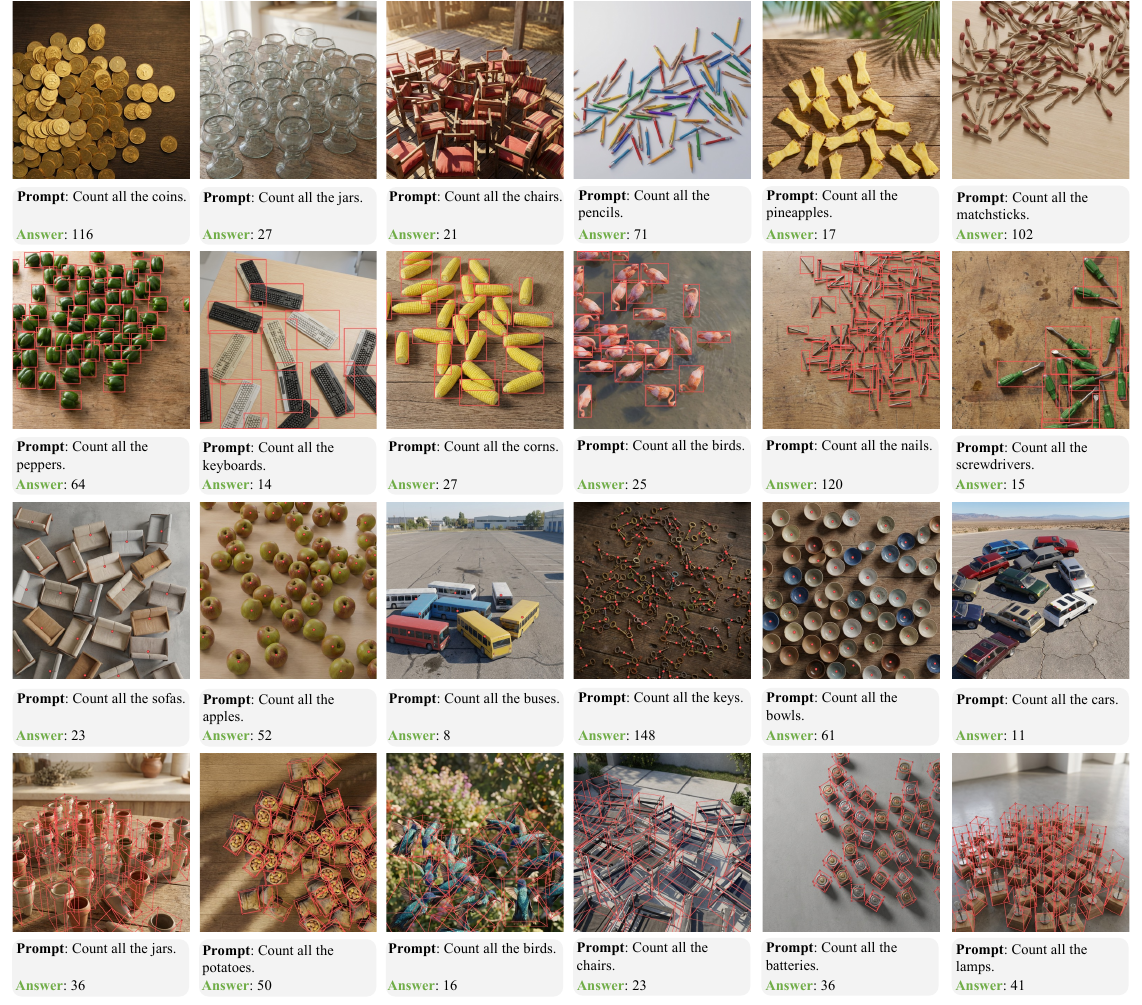}
    \vspace{-18pt}
    \caption{
        \textbf{Qualitative visualizations for Level~1.}
        Level~1 corresponds to identity-level counting, where each image contains only one object category and the task is to count all instances in the scene.
        Each example shows the scene and its corresponding counting prompt and GT answer.
    }
    \label{fig:supp_level1_visualizations}
    \vspace{-10pt}
\end{figure*}

\begin{figure*}[!htb]
    \centering
    \includegraphics[width=\textwidth]{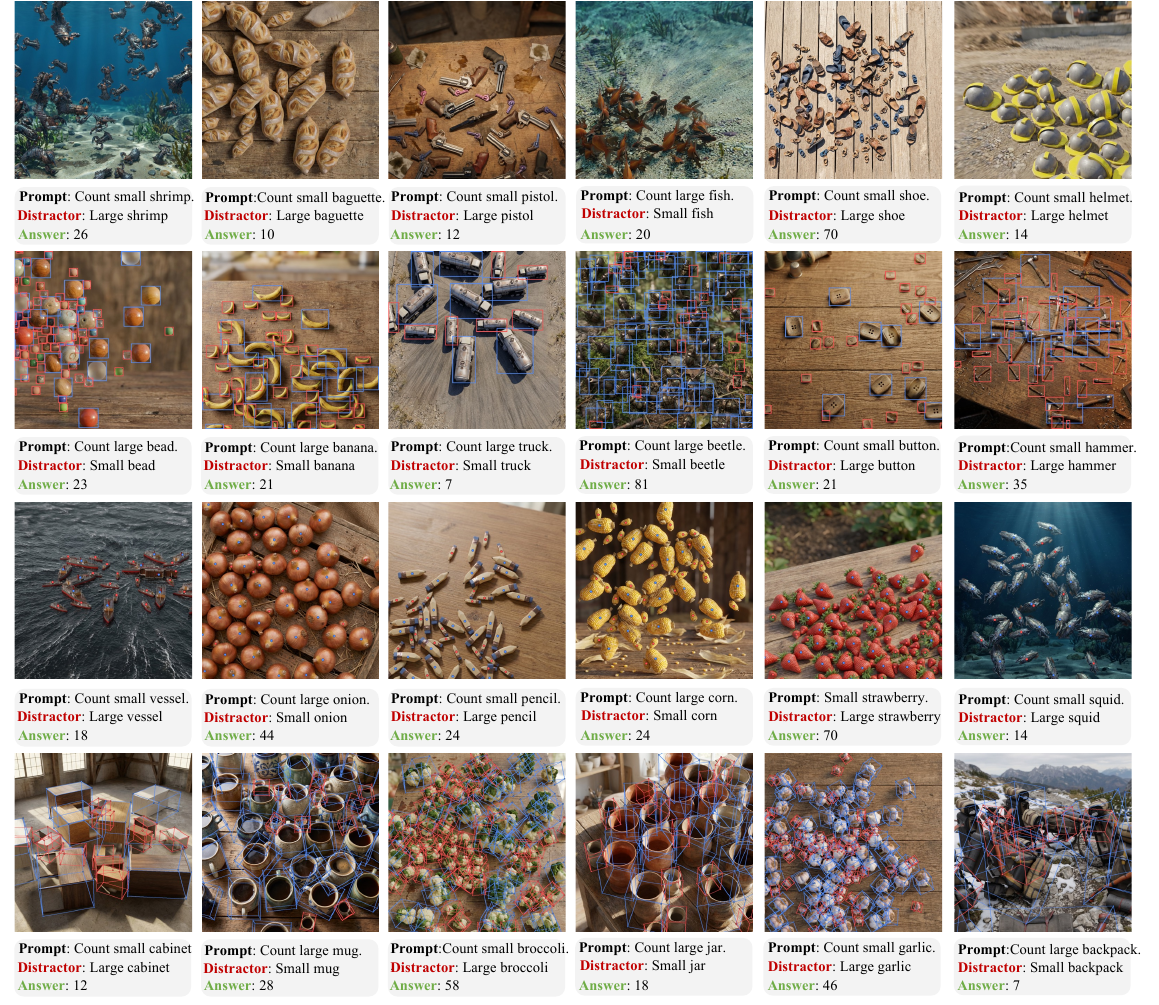}
    \vspace{-18pt}
    \caption{
        \textbf{Qualitative visualizations for Level~2~(size mode).}
        Level~2 (size mode) corresponds to attribute-level counting, where all objects belong to the same category and instance type but differ in size, requiring the model to count only the target size group.
        Each example shows the scene and its corresponding counting prompt and GT answer.
    }
    \label{fig:supp_level2_visualization_s}
    \vspace{-10pt}
\end{figure*}

\begin{figure*}[!htb]
    \centering
    \includegraphics[width=\textwidth]{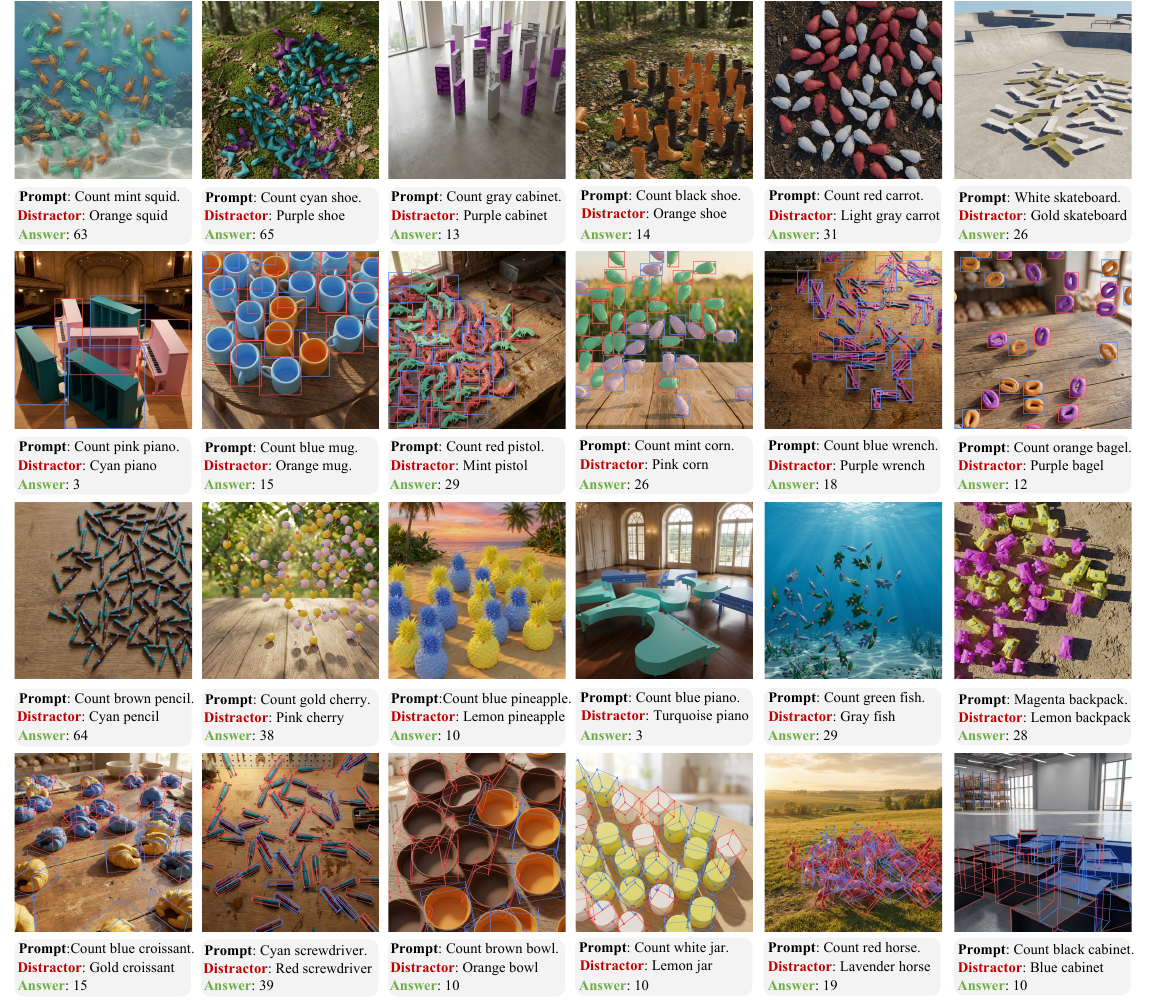}
    \vspace{-18pt}
    \caption{
        \textbf{Qualitative visualizations for Level~2~(color mode).}
        Level~2 (color mode) also corresponds to attribute-level counting, where all objects belong to the same category and instance type but differ in color, requiring the model to count only the target color group.
        Each example shows the scene and its corresponding counting prompt and GT answer.
    }
    \label{fig:supp_level2_visualization_c}
    \vspace{-10pt}
\end{figure*}

\begin{figure*}[!htb]
    \centering
    \includegraphics[width=\textwidth]{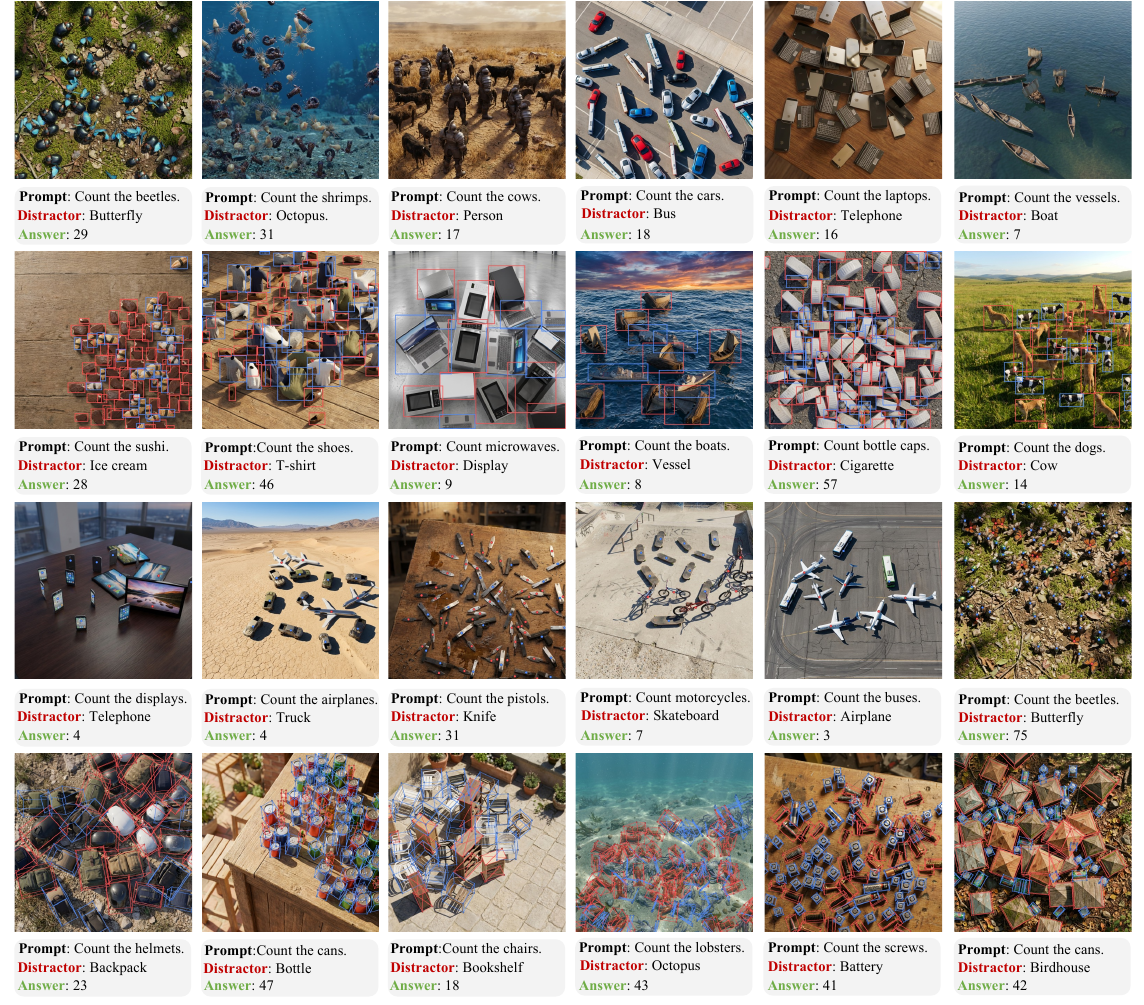}
    \vspace{-18pt}
    \caption{
        \textbf{Qualitative visualizations for Level~3.}
        Level~3 corresponds to category-level counting, where each image contains two different categories and the task is to count the target category while ignoring the distractor category.
        Each example shows the scene and its corresponding counting prompt and GT answer.
    }
    \label{fig:supp_level3_visualizations}
    \vspace{-10pt}
\end{figure*}

\begin{figure*}[!htb]
    \centering
    \includegraphics[width=\textwidth]{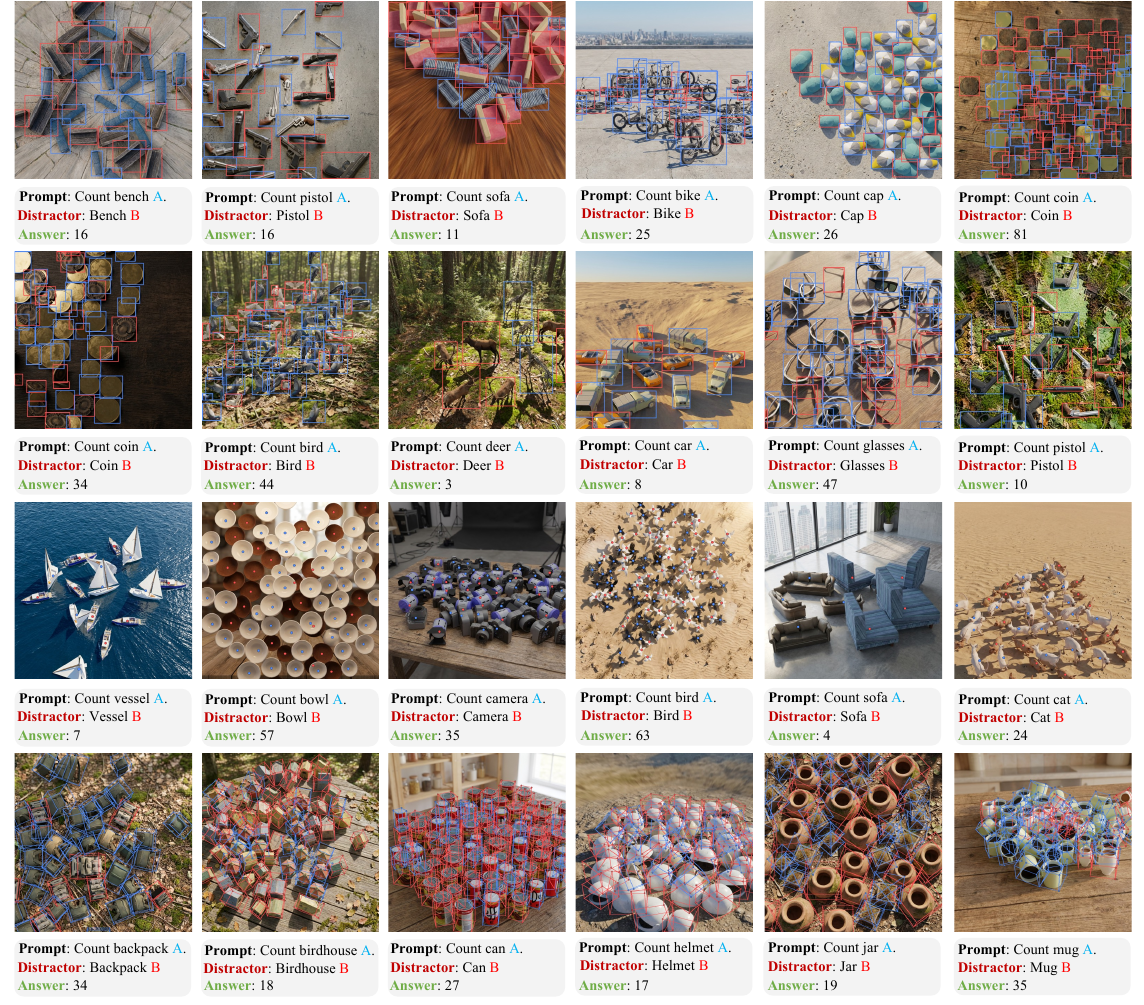}
    \vspace{-18pt}
    \caption{
        \textbf{Qualitative visualizations for Level~4.}
        Level~4 corresponds to instance-level counting, where each image contains two different instance types within the same category and the task is to distinguish and count only the target type.
        Each example shows the scene and its corresponding counting prompt and GT answer.
    }
    \label{fig:supp_level4_visualizations}
    \vspace{-10pt}
\end{figure*}

\begin{figure*}[!htb]
    \centering
    \includegraphics[width=\textwidth]{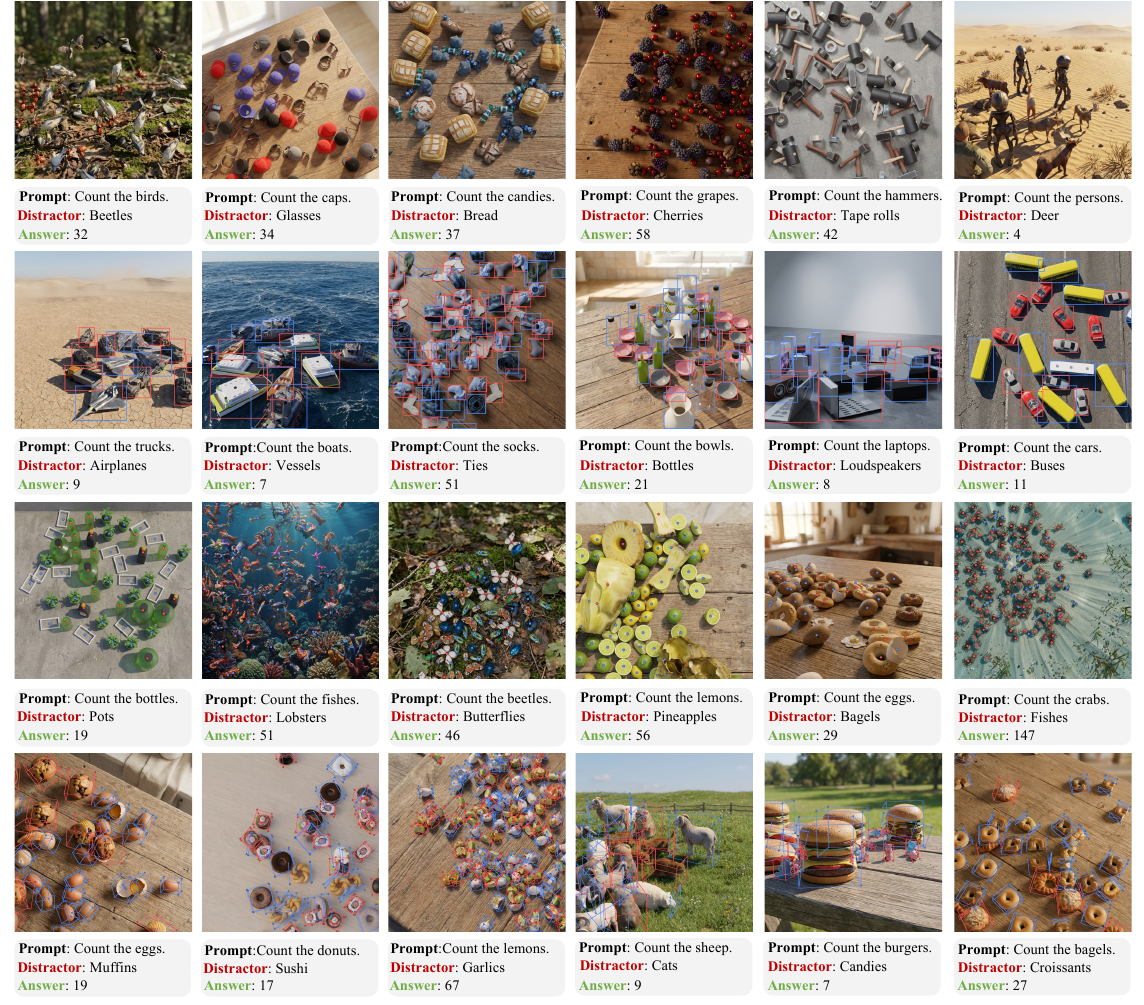}
    \vspace{-18pt}
    \caption{
        \textbf{Qualitative visualizations for Level~5.}
        Level~5 corresponds to concept-level counting, where each image contains two categories with larger intra-category variation, requiring the model to count the target category under more diverse and challenging distractor settings.
        Each example shows the scene and its corresponding counting prompt and GT answer.
    }
    \label{fig:supp_level5_visualizations}
    \vspace{-10pt}
\end{figure*}

\clearpage

\section{Additional Dataset Details}

This section provides supplementary details for the data construction process and the resulting \textbf{KubriCount} benchmark.
We first present additional implementation details of the automatic data scaling pipeline in \cref{sec:supp_scaling_pipeline_details}, including practical design choices in synthesis, editing, and filtering.
Then, we give more comprehensive dataset statistics in \cref{sec:supp_data_statistics}.

\subsection{Scaling Pipeline Details}
\label{sec:supp_scaling_pipeline_details}

Here, we provide additional implementation details of our automatic data scaling pipeline. While the main paper presents the overall design at a high level, we further describe the practical choices made in each stage of the pipeline, including 3D asset curation, prototype synthesis, consistent image editing, and automatic data filtering.

\vspace{3pt} \noindent \textbf{Stage-I: 3D asset curation.}
Here we provide additional details for the \textit{text-to-image-to-3D}.
We first identify novel categories by analyzing the coverage of existing counting datasets~\cite{fsc147,deitke2025molmo}, and then use LLMs~\cite{gpt5,gemini3} to expand each category into generation prompts.
In practice, we use two prompt formats: \textit{short captions} that directly describe a single object, and \textit{modifier+subtype descriptions} that explicitly specify a finer-grained appearance or subtype variation within the category.

An example of a \textit{short caption} prompt specification is shown below.

\begin{jsonbox}
      "bitten apple with missing chunk (single apple)"
\end{jsonbox}

An example of a \textit{modifier+subtype description} prompt specification is shown below.

\begin{jsonbox}
      "sitting upright, tail wrapped" + "maine coon cat"
\end{jsonbox}

Given these prompts, we generate single-object RGBA cutouts with Gemini-2.5-Flash-Image-Preview~\cite{nanobanana}.
To make the outputs suitable for downstream 3D reconstruction and scene composition, we require a transparent background, no cast shadows, and diverse but clean object silhouettes.
The prompt template used for RGBA cutout generation is shown below.

\begin{jsonbox}
Generate a single, photorealistic {PROMPT} as an isolated product cutout for 3D reconstruction.

Hard requirements:
- Output must be a PNG with a true transparent alpha channel (background alpha = 0 everywhere).
- No background, no floor, no shadows, no reflections, no glow, no halo, no vignette, no fog, no environment.
- Only ONE object. Fully visible in frame. No cropping. No extra items, no text, no watermark, no logo.
- Clean, sharp edges on the alpha matte; no semi-transparent fringe.

Diversity requirements (important):
- Do NOT just change color/pattern; change geometry. Avoid repeating the same silhouette across variants.
- Variant hint: {SEED}

Rendering:
- Realistic materials and lighting on the object itself, studio-quality, high detail, natural texture.
- Neutral lens perspective (no extreme wide-angle), centered, 3/4 view preferred.

Negative constraints:
- No background pixels, no drop shadow, no contact shadow, no cast shadow, no scene context.
- No multiple objects, no humans, no hands, no supports, no props.
\end{jsonbox}

We then reconstruct corresponding 3D meshes using TRELLIS.2-4B~\cite{trellis2}, which converts the generated single-object images into mesh assets that can be further normalized and imported into Kubric.

\vspace{3pt} \noindent \textbf{Stage-II: prototype synthesis.}
As noted in the main paper, scene generation is \textit{configuration-driven}, with category-specific profiles that define synthesis hyperparameters to ensure physical plausibility.
In practice, these profiles specify the valid ranges of key scene variables, such as object scale, object count, spatial density, camera pose, and placement constraints, so that different categories can be rendered under realistic yet diverse configurations.
During synthesis, the Kubric worker samples assets and scene layouts according to these profiles, runs physics simulation, and renders the resulting prototype together with exact instance-level annotations.
A configuration example is shown below.

\begin{jsonbox}
{
  "min_objects_per_group": [3, 6],
  "max_total_objects": [8, 16],
  "object_size_min": [1.00, 1.20],
  "object_size_max": [1.40, 1.70],
  "density_factor": [1.00, 1.10],
  "min_distance_ratio": [0.86, 0.94],
  "camera_distance_min": [4.8, 6.0],
  "camera_distance_max": [10.0, 13.5],
  "camera_height_min": [1.0, 2.0],
  "camera_height_max": [8.0, 12.0],
  "camera_angle_min": [10.0, 16.0],
  "camera_angle_max": [50.0, 65.0],
  "focal_length_min": [24.0, 32.0],
  "focal_length_max": [60.0, 78.0],
  "coverage_min": [0.78, 0.82],
  "coverage_max": [0.92, 0.94]
}

\end{jsonbox}

\vspace{3pt} \noindent \textbf{Stage-III: consistent image editing.}
To reduce the sim-to-real gap while preserving annotation fidelity, we use \textit{level-aware editing prompts} that explicitly specify what can and cannot be changed during image editing.
Although the overall goal is always to improve realism, the editable content depends on the counting level.
In all cases, the prompts enforce the same core constraints, namely preserving object geometry, object count, and semantic identity.
The prompt templates used for each level are shown below, with Level~2 separated into size-based and color-based variants.

The Level-1 editing prompt is shown below.
\begin{jsonbox}
Photorealistic image editing based on input RGB and a single mask for {category}.

PRIMARY DIRECTIVE: PRESERVE GEOMETRY & COUNT
Keep the object silhouettes and the number of {category} instances EXACTLY the same as indicated by the mask.

Task Instructions:
1. Object Texture Editing (Moderate Freedom, Realistic):
   - Improve realism with natural texture/material variations (e.g., subtle color variations, wear, manufacturing differences).
   - Increase intra-category diversity so instances are not identical, but keep all instances clearly {category}.
   - Avoid extreme patterns, logos, text, or implausible materials for this category.
2. Lighting & Shadows:
   - Ensure edited objects remain consistent with the global illumination (high-quality shading and plausible shadows).
3. Background Generation:
   - Generate a clean, high-quality background that is semantically coherent with {category}.
   - Keep perspective consistent (ground plane, scale, depth), and match lighting to the objects.

CRITICAL CONSTRAINTS (Strictly Adhere):
- Strict Mask Adherence: ALL edits to {category} must occur STRICTLY INSIDE the mask boundaries.
  Do NOT expand, shrink, warp, or reshape the objects. Do NOT change the silhouette.
- Zero Tolerance for New Instances: It is STRICTLY FORBIDDEN to generate any additional {category} instances in unmasked regions.
  The number of {category} objects MUST remain exactly the same as shown by the mask.
- Object Integrity: Do not remove, merge, split, or duplicate instances. Do not occlude objects with new content.
- Background Purity: Background should be purely environmental and clean. Do NOT add people, animals, or distracting objects.
- Photorealism: Output must be realistic, with consistent perspective and physically plausible lighting.
\end{jsonbox}

The Level-2 size-based editing prompt is shown below.
\begin{jsonbox}
Photorealistic image editing based on input RGB and masks for {category} (two size groups).

PRIMARY DIRECTIVE: PRESERVE SIZE DISTINCTION & COUNT
The size difference is the key signal. Preserve each instance's size and silhouette exactly as defined by the masks.

Task Instructions:
1. Object Texture Editing (Moderate Freedom, Realistic):
   - Apply realistic texture/material variations across ALL {category} instances while keeping them clearly {category}.
   - Keep textures coherent and plausible; subtle diversity is preferred over extreme style changes.
   - Ensure texture edits do not reduce perceived size difference between the two groups.
2. Background Generation:
   - Generate a coherent environment that suits {category}.
   - Maintain correct perspective, scale cues, and lighting consistency.
3. Fine Details:
   - Improve realism: consistent contact shadows, mild specular highlights, and material consistency.

CRITICAL CONSTRAINTS (Strictly Adhere):
- Strict Mask Adherence: Use the masks as rigid containers. Do NOT change silhouettes, sizes, or boundaries.
- Preserve Size Difference: Do NOT alter object geometry or perspective in a way that undermines the Large vs Small distinction.
- Zero Tolerance for New Instances: Do NOT add any new {category} instances (neither large nor small) in unmasked regions.
- Object Count Consistency: If the mask contains N instances, the output MUST contain exactly N instances.
- Background Purity: Background must be environmental only; do NOT add people/animals or distracting objects.
\end{jsonbox}

The Level-2 color-based editing prompt is shown below.
\begin{jsonbox}
Photorealistic background replacement based on input RGB and mask for {category}.

PRIMARY DIRECTIVE: PRESERVE OBJECT APPEARANCE (COLOR/TEXTURE)
The group distinction relies on original colors. You must NOT modify the object colors, textures, or materials inside the mask.

Task Instructions:
1. Objects (Strict Preservation):
   - Keep the original RGB pixels INSIDE the mask exactly unchanged.
   - Do not apply any style transfer, recoloring, or texture editing to masked regions.
2. Background Generation:
   - Replace the unmasked background with a high-quality, aesthetic environment coherent with {category}.
   - Match lighting and shadows so the objects appear naturally placed.

Additional Info:
- Group colors (do not change): {color_a_name} vs {color_b_name}

CRITICAL CONSTRAINTS (Strictly Adhere):
- No Texture/Color Changes: It is STRICTLY PROHIBITED to alter any masked pixels (color/texture/material).
- Strict Mask Adherence: Do NOT bleed edits across mask boundaries.
- Zero Tolerance for New Instances: Do NOT add extra {category} objects in the background.
- Background Purity: Keep background clean; avoid people, animals, text, or clutter.
- Photorealism: Maintain consistent perspective and lighting.
\end{jsonbox}

The Level-3 editing prompt is shown below.
\begin{jsonbox}
Photorealistic image editing based on input RGB and separate masks for {category_A} and {category_B}.

PRIMARY DIRECTIVE: CATEGORY CONSISTENCY & COUNT
Keep {category_A} and {category_B} instances unchanged in geometry and count, and keep the two categories clearly distinguishable.

Task Instructions:
1. Object Texture Editing (Moderate Freedom, Realistic):
   - Apply realistic texture/material improvements and subtle variations ONLY inside each mask.
   - Keep each instance clearly within its original category (no category drift).
   - Prefer coherent, plausible materials over extreme style changes.
2. Background Generation:
   - Generate a coherent, high-quality background that plausibly fits BOTH categories together.
   - Maintain consistent lighting direction, shadows, perspective, and scale cues.

CRITICAL CONSTRAINTS (Strictly Adhere):
- Strict Mask Adherence: Edit ONLY inside masks. Do NOT change silhouettes, boundaries, or object placement.
- ZERO TOLERANCE FOR NEW INSTANCES (Most Important):
  It is STRICTLY FORBIDDEN to generate any additional {category_A} or {category_B} objects in unmasked regions.
  The background MUST be purely environmental. Do NOT add duplicates, partial objects, reflections, posters, pictures, toys, or decorative motifs of {category_A}/{category_B}.
- Object Count Consistency: The number of {category_A} and {category_B} instances MUST remain exactly the same as indicated by the masks.
- Background Purity: Do NOT add people, animals, text, logos, or distracting objects that could be confused with the target categories.
- Photorealism: Output must be realistic and consistent with the scene geometry and illumination.
\end{jsonbox}

The Level-4 editing prompt is shown below.
\begin{jsonbox}
Photorealistic image compositing based on input RGB and separate masks for {category} Type-A and {category} Type-B.

PRIMARY DIRECTIVE: INTRA-CLASS DISTINCTION & COUNT
Type-A and Type-B must remain distinguishable. Preserve object geometry and count exactly as indicated by the masks.

Task Instructions:
1. Background (Priority):
   - Generate a coherent, photorealistic environment suitable for {category}.
   - Maintain correct perspective, geometry, and lighting.
2. Texture Editing (Conditional, Conservative):
   - You MAY apply subtle texture/material improvements to increase realism and diversity,
     ONLY IF Type-A vs Type-B remains clearly separable.
   - If there is any risk of confusion, do NOT edit object textures; keep original object appearance.

CRITICAL CONSTRAINTS (Strictly Adhere):
- Strict Mask Adherence: Do NOT change silhouettes; edit only within masks.
- Preserve Distinction (Important): Do NOT make Type-A look like Type-B or vice versa.
- Zero Tolerance for New Instances (Important): Do NOT generate any additional {category} instances in unmasked regions.
- Object Count Consistency (Important): Do NOT remove, merge, split, or duplicate objects; count must remain identical.
- Background Purity: Keep background clean and environmental; avoid people/animals and clutter.
\end{jsonbox}

The Level-5 editing prompt is shown below.
\begin{jsonbox}
Photorealistic image compositing based on input RGB and separate masks for {category_A} and {category_B}.

PRIMARY DIRECTIVE: INTER-CLASS DISTINCTION & COUNT
Ensure {category_A} and {category_B} remain clearly distinguishable. Preserve geometry and object count exactly as indicated by the masks.

Task Instructions:
1. Background (Priority):
   - Generate a coherent, high-quality environment that plausibly contains BOTH {category_A} and {category_B}.
   - Maintain consistent perspective, ground plane, and contact shadows.
2. Texture Editing (Conditional, Conservative):
   - Apply realistic texture/material improvements and mild variation ONLY IF it does not confuse categories.
   - If uncertain, keep object textures unchanged and focus on background realism.

CRITICAL CONSTRAINTS (Strictly Adhere):
- Strict Mask Adherence: Edit only within masks; do NOT change silhouettes or boundaries.
- Category Consistency (Important): Do NOT turn {category_A} into {category_B} or vice versa.
- Zero Tolerance for New Instances (Important): Do NOT add any new {category_A} or {category_B} objects in unmasked regions.
- Object Count Consistency (Important): Do NOT remove, merge, split, or duplicate any object instances. Every object indicated by the masks MUST remain present and clearly visible in the final image.
- Background Purity: Avoid people, animals, and distracting objects; keep background environmental and clean.
\end{jsonbox}

\vspace{3pt} \noindent \textbf{Stage-IV: automatic data filtering.}
After image editing, we apply a VLM-based filtering step~\cite{gemini3} to reject samples that violate the annotation-preserving constraints of the pipeline.
The inspector takes three inputs: the original RGB render, the corresponding segmentation mask(s), and the edited RGB result, and then outputs a binary \texttt{PASS}/\texttt{FAIL} decision.
The prompt is intentionally conservative: while small mask-boundary deviations are tolerated, any change in object position, object count, category identity, or the introduction of new target-category instances in background regions leads to rejection. 
The prompt also explicitly emphasizes border and corner regions, where missing-instance errors are more likely to occur.
The full filtering prompt is shown below.

\begin{jsonbox}
You are a strict visual quality inspector for a synthetic counting dataset.

You will be given:
- Image A: original RGB render
- Image B: segmentation mask(s) (one mask, or multiple masks for different groups/categories)
- Image C: edited RGB result

Your task:
Decide whether Image C is a valid edit of Image A under the constraints implied by the mask(s).
Output ONLY one token: PASS or FAIL. Do NOT output any explanation.

Acceptance note:
- Small mask-boundary/silhouette deviations are acceptable.
- Object positions, object counts, and object categories MUST remain unchanged.

PASS conditions (ALL must hold):

1) Position/layout preservation (Strict):
   - Each instance remains at the same image location as in Image A (no shifting/repositioning).
   - Global layout is unchanged.
   - No camera/viewpoint change: perspective, scale, and vanishing points remain consistent.

2) Count consistency (Strict):
   - The number of instances indicated by the mask(s) remains EXACTLY the same.
   - No instance is removed, duplicated, merged, or split.

3) Category consistency (Strict):
   - Each masked instance remains the same category as in Image A / implied by the mask(s).
   - For multi-mask (two-group/two-category) input: no swapping between masks; categories remain distinguishable.

4) Missing-instance check with edge/corner focus (Strict, with exception):
   - Focus especially on instances near image borders and corners (these are most likely to be accidentally dropped).
   - Compare Image A vs Image C at those edge/corner locations: if an instance that is present in Image A is missing in Image C, FAIL.
   - Exception: if an instance is already partially out-of-frame in Image A (cropped by the image boundary), you may ignore small visibility differences; do NOT fail solely because that already-cropped instance becomes slightly less visible. But do fail if it disappears entirely.

5) No new target-category instances, focusing on background-only regions (Strict):
   - Focus especially on regions that are background in Image A and outside the mask(s).
   - It is STRICTLY FORBIDDEN to introduce any new instances of the target categories in Image C within these originally background regions.

6) Editing locality (Moderate):
   - Object edits mainly inside masks; background edits mainly outside masks.
   - Minor boundary leakage is acceptable ONLY if it does not change position, count, or category.

7) Image integrity (Strict):
   - No severe artifacts that invalidate the sample: missing regions, duplicated edges creating extra instances, heavy blur making instances uncountable, or obvious geometric distortions.

FAIL rule:
If ANY strict check (1)-(5) or (7) fails, output FAIL.
If uncertain about any strict check, output FAIL.

Output format:
PASS
or
FAIL
\end{jsonbox}

\begin{figure*}[!htb]
    \centering
    \begin{subfigure}[t]{0.49\textwidth}
        \centering
        \includegraphics[
            width=\linewidth,
            height=5.3cm,
            keepaspectratio
        ]{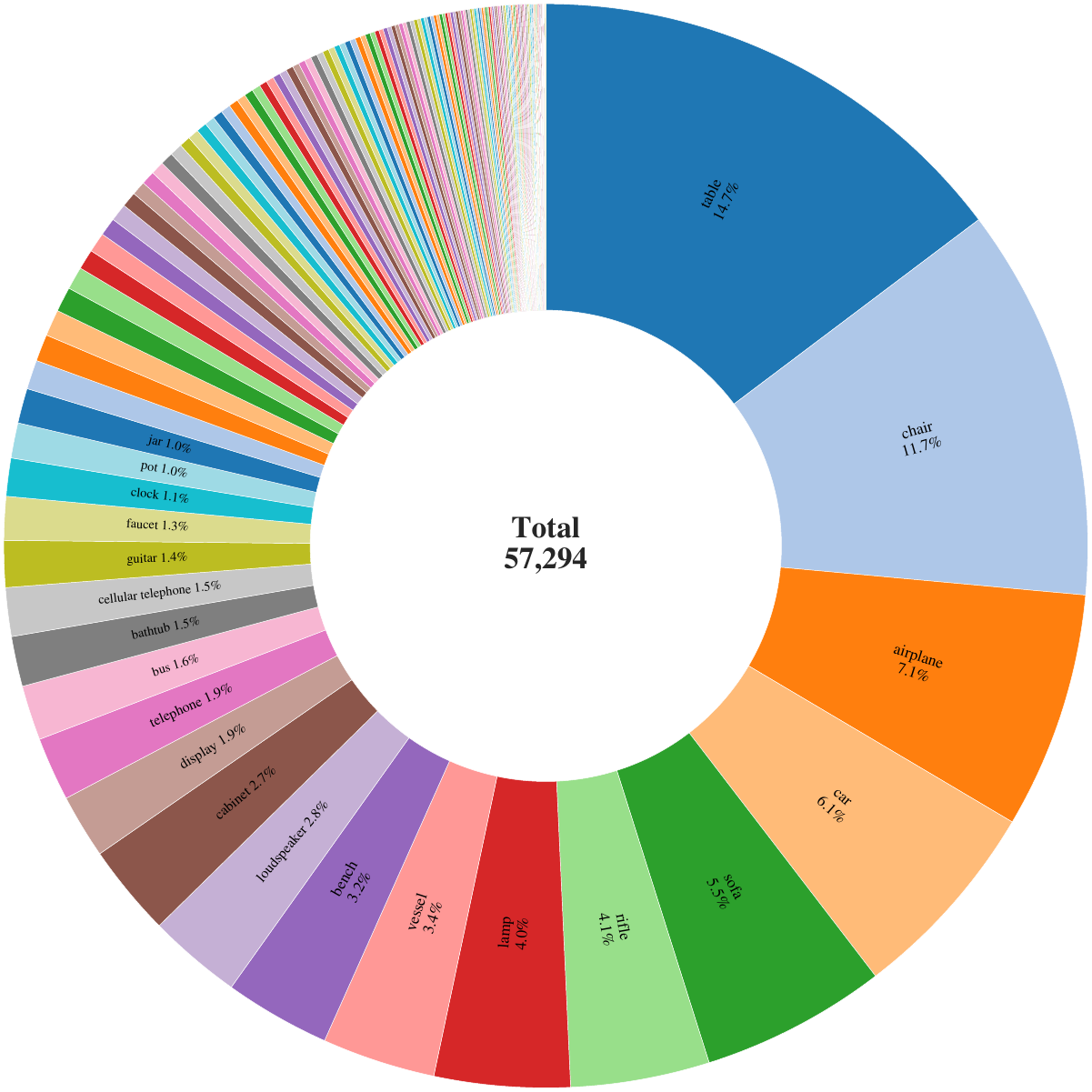}
        \caption{\textbf{3D asset category distribution.}}
        \label{fig:asset_category_distribution}
    \end{subfigure}
    \hfill
    \begin{subfigure}[t]{0.49\textwidth}
        \centering
        \includegraphics[
            width=\linewidth,
            height=5.3cm,
            keepaspectratio
        ]{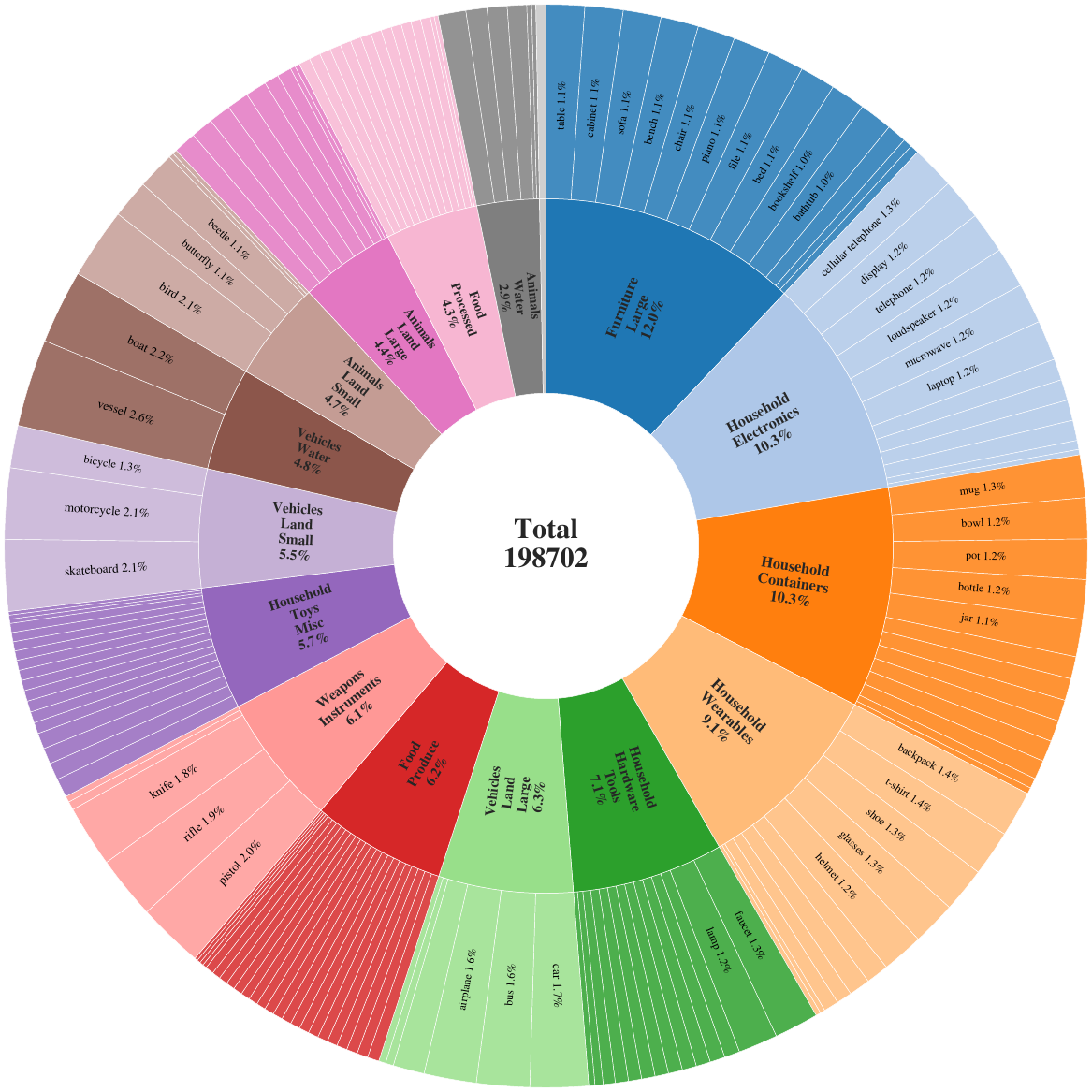}
        \caption{\textbf{Image category distribution.} }
        \label{fig:image_category_distribution}
    \end{subfigure}
    \vspace{-3pt}
    \caption{
    \textbf{Category distribution re-balance from 3D assets to generated images.}
    The curated 3D asset pool exhibits a pronounced long-tail category distribution, whereas the final image distribution is substantially more balanced after category-aware sampling during scene generation.}
    \label{fig:kubricount_statistics_supp}
    \vspace{-20pt}
\end{figure*}

\subsection{Dataset Statistics}
\label{sec:supp_data_statistics}

\vspace{3pt} \noindent \textbf{Category distribution re-balance.} The raw 3D asset pool exhibits a clear long-tail distribution over categories, as shown in \cref{fig:asset_category_distribution}.
If used directly, this imbalance would transfer to the resulting dataset and reduce category balance in the final benchmark. To mitigate this issue, we adopt a category-aware sampling strategy during scene generation: categories are sampled as uniformly as possible, while categories with very limited asset availability are assigned lower sampling probabilities to avoid excessive reuse.
The down-weighting threshold is determined by the average number of images per category.
As a result, the category distribution of the generated images becomes substantially more balanced than that of the underlying 3D asset pool, while still respecting asset availability constraints, as shown in \cref{fig:image_category_distribution}.

\input{tables/image_stat}

\vspace{3pt} \noindent \textbf{Data statistics.}
\cref{tab:supp_level_split_statistics} reports the exact number of images in each level and split.
In total, KubriCount is partitioned into Train~(99,639 images), TestA~(5,462 images featuring novel assets from seen categories), and TestB (5,406 images featuring entirely novel categories), giving 110,507 images overall.
\cref{tab:supp_super_category_statistics} and \cref{tab:supp_category_counts} further summarize the query-based image counts at the super-category and category levels, respectively. 
Since each Level-1 image yields one query, while each image in Levels~2--5 yields two queries by swapping \(\mathcal{S}^+\) and \(\mathcal{S}^-\), the dataset contains 198,702 queries in total.
For completeness, we also list the full super-category-to-category mapping in \cref{tab:supp_super_category_mapping}.

\input{tables/category_stat}

\clearpage

\section{Additional Evaluation Details}
 
This section provides supplementary details and analyses for our KubriCount evaluation.
We first describe the level-specific prompting setup used in KubriCount evaluation for MLLMs, in \cref{sec:supp_kubricount_evaluation}.
We then present further quantitative analysis in \cref{sec:supp_further_quantitative_analysis}, where we examine model predictions through prediction-versus-ground-truth scatter plots.
Finally, we show additional qualitative results in \cref{sec:supp_additional_qualitative_results} to further illustrate representative successes and failures beyond those included in the main paper.

\subsection{KubriCount Evaluation}
\label{sec:supp_kubricount_evaluation}

In this subsection, we provide the exact prompt templates used for MLLM evaluation on KubriCount.
All prompts instruct the model to directly output a single integer without additional explanation.
For Levels~2,~3, and~5, we use the same category-plus-exclusion template; for Level~4, we additionally provide one positive and one negative exemplar box in coordinate form to distinguish two instance types within the same category.

The prompt template for Level~1 is shown below.
\begin{jsonbox}
{
  "level": "Level 1",
  "prompt_template": "Please count all objects of category '{category}' in the image. Directly output the total number as an integer only. Do not output any other words. If unsure, guess a number."
}
\end{jsonbox}

The shared prompt template for Levels~2,~3, and~5 is shown below.
\begin{jsonbox}
{
  "levels": ["Level 2", "Level 3", "Level 5"],
  "prompt_template": "Please count all objects of category '{category}' in the image, and ignore objects of category '{negative_category}'. Directly output the total number as an integer only. Do not output any other words. If unsure, guess a number."
}
\end{jsonbox}

The prompt template for Level~4 is shown below.
\begin{jsonbox}
{
  "level": "Level 4",
  "prompt_template": "In the image there are two different types of objects that share the same category name '{category}'. Type A has an example bounding box {positive_box}. Type B has an example bounding box {negative_box}. Please count ONLY Type A objects and ignore Type B objects. Directly output the total number as an integer only. Do not output any other words. If unsure, guess a number."
}
\end{jsonbox}

Notably, since each Level~1 image yields a single query whereas each image in Levels~2--5 yields two queries (by swapping \(\mathcal{S}^+\) and \(\mathcal{S}^-\)), we assign Level~1 results a \textbf{2$\times$ weight} when computing the overall MAE and RMSE, so that the contribution from each level is approximately balanced in terms of query count.

\subsection{Further Quantitative Analysis}
\label{sec:supp_further_quantitative_analysis}

In this subsection, we further analyze model behavior on KubriCount through \textbf{prediction-versus-ground-truth scatter plots}.
We compare three representative models, \ie, GPT-5~\cite{gpt5}, InternVL3-78B~\cite{InternVL3}, and CountGD~\cite{countgd}, to examine their count calibration across different granularity levels.
In these plots, points closer to the diagonal indicate more accurate predictions, while larger dispersion and off-diagonal outliers reveal prediction bias, variance, and failure cases under challenging count ranges or fine-grained distractor settings.

\begin{figure*}[!htb]
    \centering
    \includegraphics[width=\textwidth]{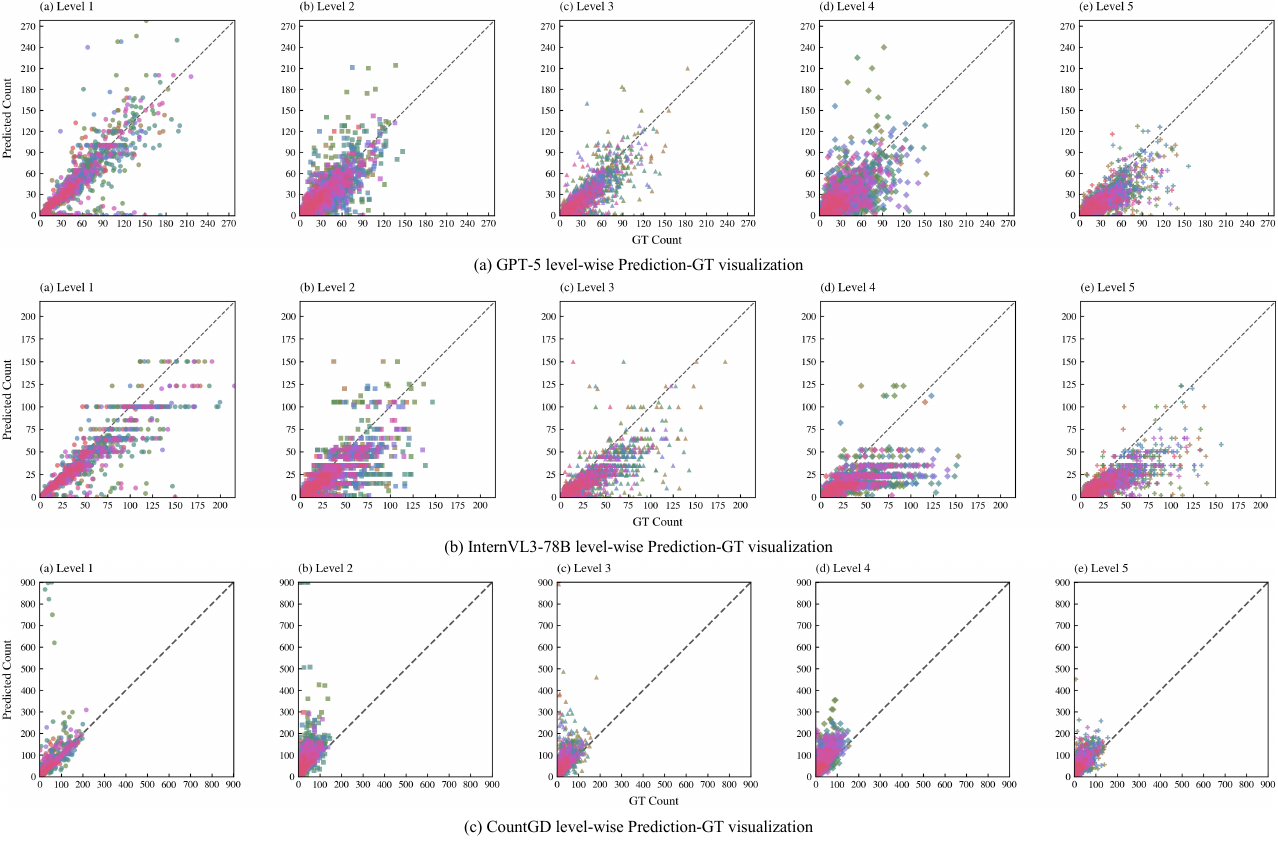}
    \vspace{-18pt}
    \caption{
        \textbf{Prediction-versus-ground-truth scatter plots on KubriCount.}
        Rows correspond to three representative models: \textbf{(a)} GPT-5, \textbf{(b)} InternVL3-78B, and \textbf{(c)} CountGD. Columns correspond to Level~1--5. Each point denotes one evaluation sample and is colored by object category. The dashed diagonal indicates perfect agreement between prediction and ground truth; deviations from this line reflect counting errors and calibration failures.
    }
    \label{fig:supp_quantitative_analysis}
    \vspace{-10pt}
\end{figure*}

From \cref{fig:supp_quantitative_analysis}, we observe distinct error patterns across model families.
GPT-5, one of the strongest proprietary MLLMs in our evaluation, shows relatively balanced behavior across levels, without an obvious level-specific or category-specific bias.
In contrast, InternVL3-78B, the strongest open-source MLLM in our evaluation, exhibits a clear systematic bias: when the ground truth count is large, its predictions tend to collapse around a few preferred values rather than tracking the ground truth smoothly.
Moreover, its errors are more often under-counting than over-counting, which becomes especially pronounced on the most challenging Level~4.
CountGD, as a positive-prompt-only counting expert model, displays another common failure pattern of specialist counting methods: occasional large outlier predictions (reaching up to 900 in Level~1), together with a clear tendency to over-count on Levels~2--5.
This behavior further supports our main finding that current counting expert models still have limited prompt-following ability in multi-category scenes with distractors.

\subsection{Additional Qualitative Results}
\label{sec:supp_additional_qualitative_results}

\begin{figure*}[t]
    \centering
    \includegraphics[width=\textwidth]{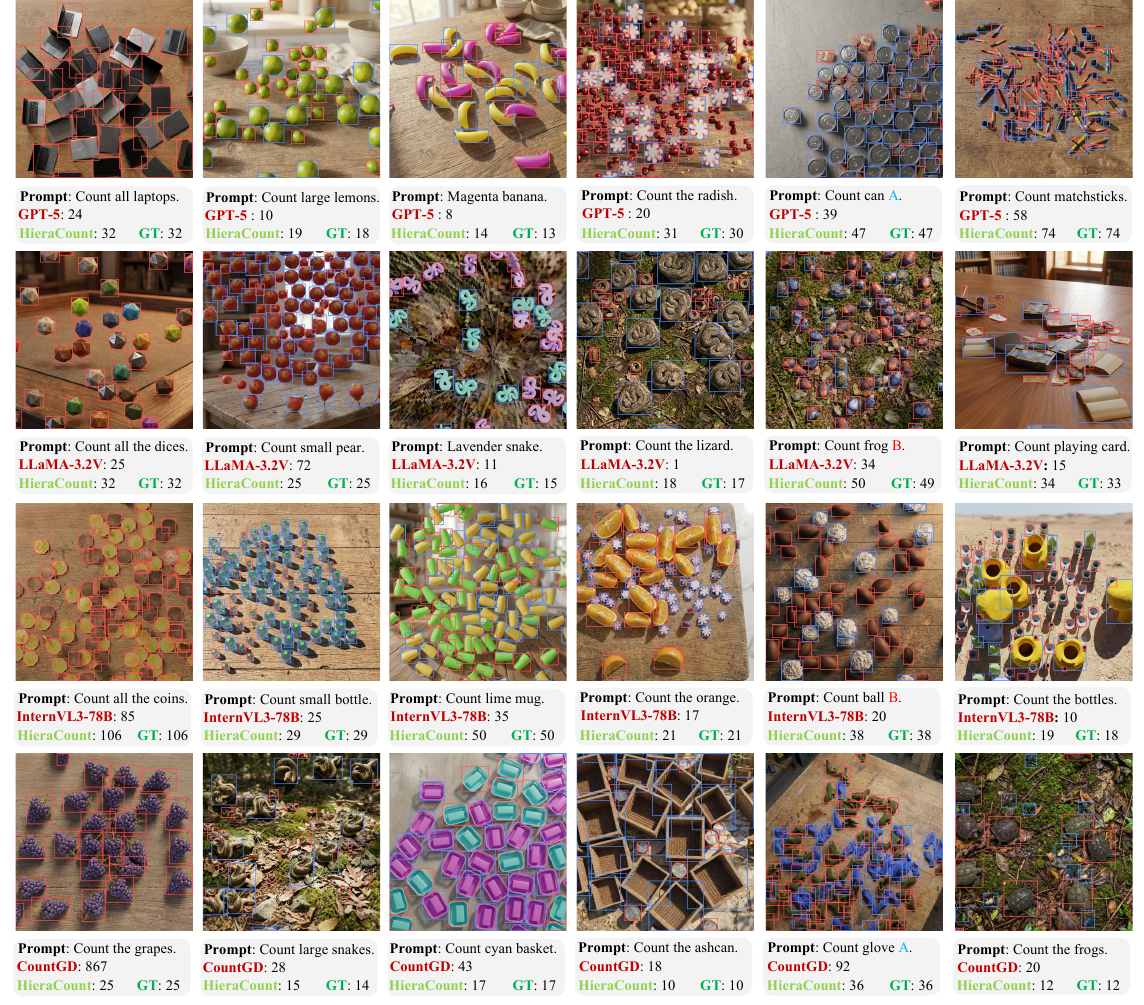}
    \vspace{-12pt}
    \caption{
        \textbf{Additional qualitative visualizations on KubriCount.}
        The examples cover diverse multi-grained queries and challenging distractor settings, highlighting common failure modes of existing models and the stronger prompt-following behavior of our HieraCount.
    }
    \label{fig:supp_additional_qualitative_results}
    \vspace{-10pt}
\end{figure*}

In this subsection, we present additional qualitative results on KubriCount to complement the examples in the main paper.
We compare HieraCount with representative strong baselines from both MLLMs and counting expert models, and include diverse prompts across different granularity levels.
These examples in~\cref{fig:supp_additional_qualitative_results} further illustrate typical prompt-following failures of existing models, as well as HieraCount's improved robustness under challenging distractor settings.

%% file: tables/image_stat.tex
\begin{table}[!htb]
    \centering
    \caption{
        \textbf{Image statistics by level and split in KubriCount.}
        Train is divided into normal and dense configurations. Level~2 is split into size and color variants.
    }
    \label{tab:supp_level_split_statistics}
    \vspace{-4pt}
    \scriptsize
    \setlength{\tabcolsep}{5pt}
    \renewcommand{\arraystretch}{1.15}
    \begin{tabular}{@{}cccccc@{}}
        \toprule
        \textbf{Level} & \textbf{Train~(Normal)} & \textbf{Train~(Dense)} & \textbf{TestA} & \textbf{TestB} & \textbf{Total} \\
        \midrule
        Level~1              & 16,179 & 3,959 & 1,087 & 1,087 & 22,312 \\
        Level~2~(Size)       & 7,582  & 2,402 & 569   & 586   & 11,139 \\
        Level~2~(Color)      & 8,043  & 2,135 & 600   & 602   & 11,380 \\
        Level~3              & 15,386 & 3,624 & 1,053 & 1,014 & 21,077 \\
        Level~4              & 16,493 & 4,186 & 1,081 & 1,081 & 22,841 \\
        Level~5              & 15,825 & 3,825 & 1,072 & 1,036 & 21,758 \\
        \midrule
        \textbf{Total}       & \textbf{79,508} & \textbf{20,131} & \textbf{5,462} & \textbf{5,406} & \textbf{110,507} \\
        \bottomrule
    \end{tabular}
    \vspace{-6pt}
\end{table}

\begin{table}[!htb]
    \centering
    \caption{
        \textbf{Super-category statistics of KubriCount.}
        \textbf{\#Cat} denotes the number of categories in each super-category, and \textbf{\#Img} denotes the query-based image count.
    }
    \label{tab:supp_super_category_statistics}
    \vspace{-4pt}
    \scriptsize
    \setlength{\tabcolsep}{6pt}
    \renewcommand{\arraystretch}{1.12}
    \begin{tabular}{lccc}
        \toprule
        \textbf{Super-category} & \textbf{Train Cat.} & \textbf{Test Cat.} & \textbf{\#Img} \\
        \midrule
        Vehicles\_Water & 2 & 0 & 9,562 \\
        Vehicles\_Land\_Large & 4 & 2 & 12,431 \\
        Vehicles\_Land\_Small & 3 & 0 & 10,942 \\
        Animals\_Water & 4 & 2 & 5,757 \\
        Animals\_Land\_Large & 7 & 2 & 8,811 \\
        Animals\_Land\_Small & 3 & 2 & 9,242 \\
        Food\_Produce & 20 & 4 & 12,378 \\
        Food\_Processed & 13 & 2 & 8,605 \\
        Furniture\_Large & 11 & 2 & 23,916 \\
        Household\_Electronics & 10 & 2 & 20,402 \\
        Weapons\_Instruments & 3 & 2 & 12,160 \\
        Household\_Containers & 11 & 2 & 20,379 \\
        Household\_Wearables & 9 & 2 & 18,134 \\
        Household\_Hardware\_Tools & 12 & 2 & 14,019 \\
        Household\_Toys\_Misc & 15 & 3 & 11,370 \\
        Structures & 1 & 0 & 594 \\
        \midrule
        \textbf{Total} & \textbf{130} & \textbf{27} & \textbf{198,702} \\
        \bottomrule
    \end{tabular}
    \vspace{-12pt}
\end{table}

%% file: tables/category_stat.tex
\begin{table*}[!htb]
    \centering
    \caption{
        \textbf{Category-level image counts in KubriCount.}
        We report query-based image counts for all 157 categories, sorted in descending order and arranged column-wise.Some long category names are abbreviated for compactness.
    }
    \label{tab:supp_category_counts}
    \vspace{-4pt}
    \scriptsize
    \setlength{\tabcolsep}{2.5pt}
    \renewcommand{\arraystretch}{0.95}
    \begin{tabular}{@{}
        >{\raggedright\arraybackslash}p{1.78cm}r
        >{\raggedright\arraybackslash}p{1.78cm}r
        >{\raggedright\arraybackslash}p{1.78cm}r
        >{\raggedright\arraybackslash}p{1.78cm}r
    @{}}
        \toprule
        \textbf{Category} & \textbf{\#Img} &
        \textbf{Category} & \textbf{\#Img} &
        \textbf{Category} & \textbf{\#Img} &
        \textbf{Category} & \textbf{\#Img} \\
        \midrule
        vessel & 5,188 & bookshelf & 2,063 & saw & 779 & bead & 543 \\
        boat & 4,374 & bathtub & 1,959 & pliers & 775 & garlic & 541 \\
        skateboard & 4,234 & truck & 1,857 & wrench & 714 & candle & 541 \\
        bird & 4,196 & fish & 1,649 & pineapple & 694 & stove & 539 \\
        motorcycle & 4,196 & cap & 1,342 & matchstick & 670 & tomatoes & 537 \\
        pistol & 3,909 & horse & 1,331 & sushi & 668 & orange & 537 \\
        rifle & 3,731 & deer & 1,324 & croissant & 665 & printer & 530 \\
        knife & 3,656 & birdhouse & 1,322 & teddy bear & 660 & broccoli/caulif. & 523 \\
        car & 3,448 & person & 1,308 & fork & 660 & grape & 516 \\
        bus & 3,142 & dog & 1,304 & egg & 657 & washer & 511 \\
        airplane & 3,097 & cup & 1,301 & carrot & 648 & guitar & 507 \\
        backpack & 2,719 & can & 1,265 & muffin & 646 & potatoes & 498 \\
        T-shirt & 2,707 & camera & 1,262 & cucumber & 645 & onion & 496 \\
        shoe & 2,655 & plate & 1,259 & pretzel & 641 & clock & 482 \\
        faucet & 2,625 & cat & 1,256 & cigarette & 639 & train & 472 \\
        glasses & 2,603 & trousers & 1,256 & sandwich & 634 & rocket & 415 \\
        cell phone & 2,563 & bag & 1,249 & ice cream & 631 & basket & 375 \\
        mug & 2,526 & octopus/squid & 1,242 & burger & 629 & pillow & 365 \\
        bicycle & 2,512 & earphone & 1,240 & pepper veg. & 620 & bat & 357 \\
        helmet & 2,474 & remote ctrl. & 1,216 & toilet paper & 618 & microphone & 318 \\
        display & 2,457 & keyboard & 1,207 & candy & 614 & elephant & 297 \\
        telephone & 2,452 & crab & 1,189 & strawberry & 613 & hat & 291 \\
        bowl & 2,418 & dishwasher & 1,167 & bagel & 609 & glove & 286 \\
        loudspeaker & 2,395 & lobster/shrimp & 1,147 & key & 607 & bear & 284 \\
        microwave & 2,388 & mailbox & 1,129 & cherry & 607 & turtle & 284 \\
        pot & 2,385 & pencil & 1,116 & apple & 600 & snake & 283 \\
        laptop & 2,374 & spoon & 1,018 & ashcan & 600 & lizard & 263 \\
        lamp & 2,343 & ball & 947 & bottle cap & 599 & dice & 251 \\
        bottle & 2,331 & coin & 943 & tower & 594 & frog & 246 \\
        table & 2,279 & sock & 916 & lemon/lime & 588 & peach & 243 \\
        butterfly & 2,269 & paperclip & 912 & donut & 585 & playing card & 242 \\
        cabinet & 2,256 & nail & 893 & watermelon & 584 & cake & 237 \\
        sofa & 2,248 & tie & 885 & baguette & 582 & pizza & 231 \\
        bench & 2,240 & cow & 862 & radish & 581 & pumpkin & 220 \\
        beetle & 2,231 & tape roll & 849 & eggplant & 579 & book & 210 \\
        jar & 2,219 & sheep & 845 & bread loaf & 576 & avocado & 207 \\
        chair & 2,218 & battery & 844 & banana & 570 & pear & 179 \\
        piano & 2,191 & hammer & 838 & button & 560 &  &  \\
        file & 2,145 & screw & 800 & corn & 552 &  &  \\
        bed & 2,100 & screwdriver & 800 & lego brick & 546 &  &  \\
        \midrule
        \multicolumn{8}{c}{\textbf{Total: 157 categories, 198,702 queries}} \\
        \bottomrule
    \end{tabular}
    \vspace{-6pt}
\end{table*}

\begin{table*}[!htb]
    \centering
    \caption{
        \textbf{Super-category to category mapping in KubriCount.}
        We list the categories used for training and those reserved for test-only (unseen categories).
    }
    \label{tab:supp_super_category_mapping}
    \vspace{-4pt}
    \scriptsize
    \setlength{\tabcolsep}{4pt}
    \renewcommand{\arraystretch}{1.12}
    \begin{tabularx}{\textwidth}{@{}lXX@{}}
        \toprule
        \textbf{Super-category} & \textbf{Train categories} & \textbf{Test-only categories} \\
        \midrule
        Vehicles\_Water & vessel; boat & -- \\
        Vehicles\_Land\_Large & airplane; car; bus; truck & train; rocket \\
        Vehicles\_Land\_Small & motorcycle; bicycle; skateboard & -- \\
        Animals\_Water & fish; octopus squid; crab; lobster shrimp & turtle; frog \\
        Animals\_Land\_Large & cat; dog; horse; deer; cow; sheep; person & elephant; bear \\
        Animals\_Land\_Small & bird; butterfly; beetle & lizard; snake \\
        Food\_Produce & banana; grape; apple; strawberry; tomatoes; orange; potatoes; carrot; onion; lemon lime; cucumber; eggplant; pepper vegetable; broccoli cauliflower; radish; garlic; corn; watermelon; pineapple; cherry & pear; avocado; pumpkin squash; peach \\
        Food\_Processed & burger; donut; sandwich; baguette; bread loaf; croissant; muffin; bagel; pretzel; candy; egg; sushi; ice cream & cake; pizza \\
        Furniture\_Large & table; chair; sofa; bench; cabinet; bookshelf; bed; piano; file; bathtub; dishwasher & stove; washer \\
        Household\_Electronics & laptop; computer keyboard; microwave; telephone; cellular telephone; loudspeaker; camera; remote control; earphone; display & printer; microphone \\
        Weapons\_Instruments & rifle; pistol; knife & guitar; bat \\
        Household\_Containers & pot; jar; bottle; mug; bowl; can; cup; plate; bag; mailbox; birdhouse & ashcan; basket \\
        Household\_Wearables & shoe; T-shirt; trousers; glasses; cap; backpack; tie; sock; helmet & hat; glove \\
        Household\_Hardware\_Tools & faucet; lamp; hammer; pliers; screwdriver; wrench; saw; nail; screw; paper clip; tape roll; battery & pillow; clock \\
        Household\_Toys\_Misc & teddy bear; ball; lego brick; coin; bottle cap; bead; button; toilet paper; pencil; fork; key; spoon; candle; matchstick; cigarette & dice; playing card; book \\
        Structures & tower & -- \\
        \bottomrule
    \end{tabularx}
    \vspace{-6pt}
\end{table*}